\documentclass[a4paper]{article}

\usepackage[pages=all, color=black, position={current page.south}, placement=bottom, scale=1, opacity=1, vshift=5mm]{background}
\SetBgContents{
	\tt This work is shared under a \href{https://creativecommons.org/licenses/by-sa/4.0/}{CC BY-SA 4.0 license} unless otherwise noted
}      

\usepackage[margin=1in]{geometry} 

\usepackage{amsmath}
\usepackage{amsthm}
\usepackage{amssymb}
\usepackage{bm}
\usepackage[utf8]{inputenc}
\usepackage{hyperref}
\usepackage{here}
\usepackage{url}
\hypersetup{
	unicode,
	pdfauthor={Author One, Author Two, Author Three},
	pdftitle={A simple article template},
	pdfsubject={A simple article template},
	pdfkeywords={article, template, simple},
	pdfproducer={LaTeX},
	pdfcreator={pdflatex}
}

\newif\ifarxiv
\arxivtrue

\theoremstyle{plain}

\theoremstyle{definition}

\usepackage{graphicx, color}
\graphicspath{{fig/}}

\usepackage{algorithm, algpseudocode} 
\usepackage{mathrsfs} 
\usepackage{natbib}
\usepackage{hyperref}
\usepackage{setspace}
\usepackage{tipa}

\title{Offensive Lineup Analysis in Basketball with Clustering Players Based on Shooting Style and Offensive Role} 
\ifarxiv

\author{Kazuhiro Yamada$^1$ \and Keisuke Fujii$^{1,2,3}$\thanks{Corresponding author: Keisuke Fujii, Nagoya University, Furocho 1, Nagoya, 464-8603, JAPAN.
\\ $+$81-52-789-3626, \texttt{fujii@i.nagoya-u.ac.jp} \\
This work was supported by JSPS KAKENHI (Grant Numbers 20H04075 and 21H05300) and JST Presto (Grant Number JPMJPR20CA).}}
\date{
	$^1$Graduate School of Informatics, Nagoya University \\ %
	$^2$RIKEN Center for Advanced Intelligence Project\\
	$^3$ Japan Science and Technology Agency
	\\[2ex]%
}
\else
\author{anonymous}
\fi 

\begin{document}
\maketitle 
\begin{abstract}
In a basketball game, scoring efficiency holds significant importance due to the numerous offensive possessions per game. Enhancing scoring efficiency necessitates effective collaboration among players with diverse playing styles. In previous studies, basketball lineups have been analyzed, but their playing style compatibility has not been quantitatively examined. The purpose of this study is to analyze more specifically the impact of playing style compatibility on scoring efficiency, focusing only on offense. This study employs two methods to capture the playing styles of players on offense: shooting style clustering using tracking data, and offensive role clustering based on annotated playtypes and advanced statistics. For the former, interpretable hand-crafted shot features and Wasserstein distances between shooting style distributions were utilized. For the latter, soft clustering was applied to playtype data for the first time. Subsequently, based on the lineup information derived from these two clusterings, machine learning models Bayesian models that predict statistics representing scoring efficiency were trained and interpreted. These approaches provide insights into which combinations of five players tend to be effective and which combinations of two players tend to produce good effects.

\end{abstract}

	\newpage
	
\section{Introduction}
In a basketball game, scoring efficiency is crucial to winning because possessions are repeated over and over again. In the United States' National Basketball Association (NBA), the average number of team possessions for the 2012-13 season was less than 100 for all teams; however, for the 2022-23 season, 28 out of 30 teams had more than 100 possessions \footnote{\url{https://www.teamrankings.com/nba/stat/possessions-per-game}}, highlighting the increasing importance of scoring efficiency.
Increasing scoring efficiency in basketball simply means consistently taking shots with high-scoring expectations. To achieve this, players with different playing styles need to collaborate effectively. Therefore, analyzing the compatibility between players can offer insights that contribute to winning the game.

When analyzing player compatibility using data, regression analysis has been conducted using the performance of a particular lineup as the objective variable. There are two primary approaches to selecting explanatory variables. One is to assess player compatibility by using the presence or absence of a particular player as explanatory variables. For example, \cite{ishida2023bayesian} conducted Bayesian modeling by including a flag indicating the presence of two specific players on offense as explanatory variables, and estimated their regression coefficient on scoring efficiency as duo effects. The other direction is to examine the compatibility between clusters (i.e., general compatibility of playing styles) by defining the playing styles of players through clustering and then using the number of players in each cluster in that lineup as an explanatory variable. \cite{kalman2020nba} redefined player positions (playing styles) by clustering players using statistics related to the frequency of play and success rate of shots, known as stats, as player features.  They then examined which cluster of players in the lineup would simultaneously increase efficiency in terms of high point-gain/low point-loss, based on the results of the clustering. However, because the features used for clustering in this study included both offensive and defensive features simultaneously, it was not possible to specifically analyze which aspect of offense or defense is more compatible with each other. As a result, the findings obtained were considered limited. For improved interpretability, focusing solely on offensive features from the beginning would be advantageous. In addition, when examining the compatibility between clusters, there remained the issue of the lack of quantitative analysis because less interpretable machine learning models were used.
Therefore, this study solely focuses on offense and aims to estimate playing style compatibility quantitatively and to provide general insight into the compatibility.

\cite{alagappan2012from} redefined player positions (playing styles) by clustering players based on offensive and defensive information for the first time. \cite{bianchi2017role} utilized basic statistics such as points, rebounds, blocks, assists, steals, turnovers, personal fouls, and points per game for clustering. \cite{kalman2020nba} utilized 23 varied metrics, such as player height and dunk rate, to define nine new positions using Gaussian Mixture Model (GMM)-based clustering.
Moreover, \cite{muniz2022weighted} performed clustering using the six types of data: points, passes, rebounds, defense, hustle, and clutches, and then analyzed the results using weighted network clustering to comprehensively cluster players. 

Clustering with offensive information only alone aims to specifically capture the player's playing style on offense. For instance, one study separates offense and defense and then makes a categorization \citep{hua2023estimating}, and other studies clustered shooting styles based on players' shot charts \citep{Fan2023} and on a specific position (e.g., guards) based on stats \citep{zhang2016application}.  However, none of the approaches included dynamic information such as how the player moved before the shot. 
Although \cite{Chen2023modeling} clustered players based on their tendency towards offensive playtypes and analyzed the impact of each cluster on team performance through logistic regression, they did not examine the compatibility between clusters.

This study introduces new approaches for analyzing basketball offensive lineups using two clustering methods for players: one based on shooting style and the other on offensive role. The former employs a hard clustering method that reflects more specific player shooting styles than the method of \cite{Fan2023} by using tracking data and including information at and just before the time of shooting. The latter method involves soft clustering players based on the percentage of playtypes and two offensive statistics.

Methodologically, various movement summarization and clustering methods are considered. 
Previous studies on dimensionality reduction of basketball players' complex movements have applied techniques such as non-negative matrix factorization \citep{Miller14}, topic modeling \citep{Miller17}, and tensor decomposition \citep{Papalexakis18}.
To capture more dynamic aspects of data, approaches like neural network-based image processing \citep{Wang16,Nistala18} and dynamic mode decomposition \citep{Fujii17, Fujii18, Fujii20} have been explored. However, in this study, we use interpretable hand-crafted features (e.g., \cite{ Mcqueen14, Hojo18, Mcintyre16, Hojo19,zhang2022cooperative}, to interpret the results of clustering in shooting styles. 

In clustering, where researchers aim to organize similar objects into groups in sports data, various similarity measures have been employed. These include dynamic time warping \citep{Sha16}, Frechet distance \citep{Kanda20}, Hausdorff distance \citep{bunker2023}, Gaussian mixture model clustering \citep{Pervse09}, and self-organizing maps \citep{Kempe2014}. However, in shooting style clustering, it is necessary to calculate the dissimilarity between distributions of the players' shots. To solve this issue, we use the Wasserstein distance \citep{Wasserstein}, which is a distance function between two distributions with fewer restrictions (for details, see Methods). 
For the latter method, we apply soft clustering on annotated playtype data provided by Synergy Sports \citep{playtype}. 
Moreover, a new method for analyzing lineups represented by clusters was proposed. Specifically, it is a Bayesian estimation of the effect of the number of combinations of two players' playing styles (clusters including results of soft clustering) in the lineup as an explanatory variable. This differs from the method of \cite{kalman2020nba} in that we quantitatively estimate the effect of the combination of playing style, allowing for comparisons of large and small effects.


The main contributions of this work are as follows: (i) We proposed a method that clusters players' shooting styles based on the Wasserstein distance, which measures the distance between distributions, using dynamic information from tracking data. 
(ii) We also proposed a method that uses playtype data and advanced stats for soft clustering to determine a player's offensive role.
(iii) We proposed a method to quantitatively analyze the compatibility between clusters (playing styles).
(iv) We provided new insights into the compatibility of playing styles in offenses such as those of ball-handlers and shooters.

\section{Methods}
Our methods consist of shooting style clustering, offensive role clustering, and lineup analysis. 
As these components are independent, we separately describe the datasets we used, preprocessing, clustering procedures, variable creation for lineup analysis, and procedures of lineup analysis. 

\subsection{Dataset}
\subsubsection{For Shooting Style Clustering}
For clustering based on shooting style, we used tracking data from SportVU, a video-based tracking system from STATS LLC (currently Stats Perform). This dataset comprises attack segments, recorded from the moment the ball crosses the center line or is brought into the frontcourt for a throw-in until the end of the possession. The data consists of a time series of player and ball location coordinates. It is important to note that the sampling rate of the original data was 25 Hz but was downsampled to 10 Hz, and all offensive directions were unified. The dataset used in this study was offense segments from 630 games within the 2015-16 NBA season. A total of 41,160 shots (28,427 two-pointers and 12,733 three-pointers) were included in the analysis, limited to attack segments that lasted more than 3 seconds and ended with a shot. A time interval from 3 seconds before the shot to the moment of the shot was analyzed. 

\subsubsection{For Offensive Role Clustering}
For clustering based on offensive role, we used playtype data obtained from NBA.com. Playtypes refer to 11 different forms of offense as defined and annotated by Synergy Sports, indicating how the possession was used. When a player performs a specific playtype of offense and the possession ends by either the player shooting or committing a turnover, the player is attributed one count for that playtype. Table \ref{tab:playtype} shows the playtypes and their brief descriptions. For further details refer to \citep{playtype}. Playtype data has been available since the 2015-16 season, and we utilized data from the 2015-16 to 2022-23 seasons in this study.  In addition, we extracted advanced stats of players in the same seasons to serve as features of players. In this study, the same player across different seasons was treated as a different distinct record, following the approach of \cite{kalman2020nba}.

\begin{table}[H]
\caption{Playtypes and the descriptions.}
\label{tab:playtype}
\centering
\scalebox{0.88}{
\begin{tabular}{lllllllllllllll}
\hline
Playtype & Description \\
\hline
Pick-and-roll ball-handler & Possessions where the player uses an on-ball screen (including reject action). \\
Pick-and-roll roll man & Possessions where the player makes a pick and then rolls, pops, slips, etc. \\
Transition & Possessions where the player attacks when defense is not set. \\
Off-screen & Possessions where the player uses an off-ball screen. \\
Spot-up & Possessions where the player stands still or moves without using an off-ball screen. \\
Isolation & Possessions where the player performs 1on1. \\
Hand-off & Possessions where the player receives the ball in a hand-off. \\
Cut & Possessions where the player cuts without a screen (including UCLA, flex, etc.). \\
Putback & Possessions where the player shoots immediately after an offensive rebound. \\
Post-up & Possessions where the player performs post-play. \\
Miscellaneous & Possessions of offense in a form that does not fit any of the above. \\
\hline
\end{tabular}
}
\end{table}

\subsubsection{For Lineup Analysis}
Data for the lineup analysis was obtained from NBA.com. In this study, offensive rating (OFFRTG) was used as a stat for lineup scoring efficiency. OFFRTG represents points per 100 possessions. For lineup analysis using the clustering results from the proposed method, we used the advanced stats of lineups from the 2012-13 to 2018-19 seasons, and for the analysis using the clustering results of the other method, we used data from the 2015-16 to 2022-23 seasons.

\subsection{Preprocessing} \label{sec:preprocessing}
\subsubsection{For Shooting Style Clustering}
\begin{figure}[htbp]
  \begin{minipage}[b]{0.48\linewidth}
    \centering
    \includegraphics[keepaspectratio, scale=0.52]{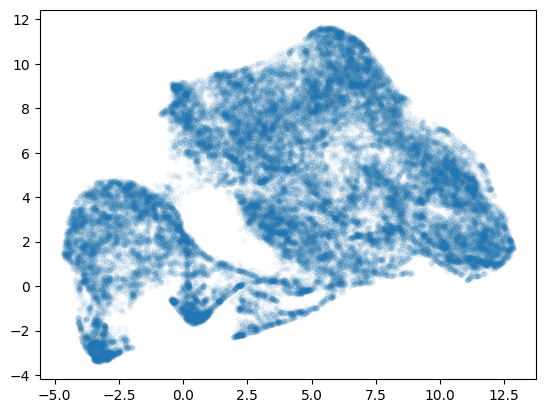}
  \end{minipage}
  \begin{minipage}[b]{0.48\linewidth}
    \centering
    \includegraphics[keepaspectratio, scale=0.52]{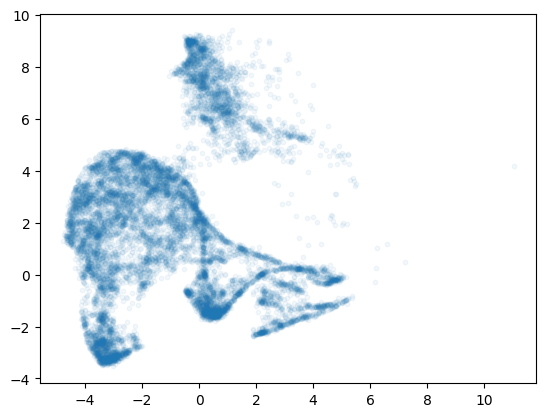}
  \end{minipage}
  \caption{Scatterplot on the feature space of shots whose features are reduced to two dimensions by UMAP; left: all shots, right: 3-pointers.}
  \label{fig:umap}
\end{figure}

We used 17 features of the shot for the subsequent analysis: x,y coordinates of the shooter at the time of the shot and 1 second before the shot [m]; x,y coordinates at the time of receiving the ball [m]; distance to the rim at the time of the shot and 0.5 seconds, 1 second, 1.5 seconds, 2 seconds, 2.5 seconds, and 3 seconds before the shot [m]; distance to the rim when the ball was received [m]; distance traveled while holding the ball [m]; speed at the time of the shot [m/s]; and time of holding the ball [s].
The features of the position coordinates were standardized due to their larger scale compared to others, and their importance is lower than that of the distances. In addition, to reduce redundancies they were compressed to 9 dimensions using principal component analysis, which retained 99\% of the cumulative contribution ratio.
For visual confirmation, we further compressed them into two dimensions using UMAP \citep{UMAP} and plotted them in Figure \ref{fig:umap}. UMAP is a non-linear dimensionality reduction method to visualize the structure of high-dimensional data and has the advantage of processing large amounts of data faster than t-SNE \citep{t-SNE}, which is a well-known non-linear dimensionality reduction method. The plot revealed a bias of three-pointers towards the left, indicating their proximity in the feature space.

\subsubsection{For Offensive Role Clustering}
 The percentage of each playtype was used as the main player's feature. Note that there are missing values because the data are recorded only for the playtypes of players who played at least 10 minutes per game in the season and at least 10 possessions. The missing values were imputed in the same way as \cite{hua2023estimating}, by equally assigning the unidentified percentages. In addition, for the purpose of scale adjustment, the percentages of Off-screen and Hand-off, which are commonly performed by movement shooters, were combined. As a constraint, we excluded player data for which the percentage of playtype was not known for more than 50\% of the players. Furthermore, two additional offensive stats, AST\% (assist percentage) and USG\% (usage percentage), were added to the player's features. AST\% represents the percentage of assists by a player on the court relative to the total number of successful shots by the rest of the players on the court, while USG\% represents the percentage of possessions ended by that player's offense while he was on the court. To ensure the reliability of these stats, we further limited them to players who played at least 20 games per season. Ultimately, 3051 players were included in the analysis.

\subsection{Clustering}
\subsubsection{Shooting Style Clustering}
The method we propose in this study involves the following steps. First, we calculated the dissimilarity of shot tendencies between players based on the Wasserstein distance. Next, hierarchical clustering was performed to group the players based on the dissimilarity matrix.

First, we calculated the dissimilarity (i.e., distance) between the distributions of the players' shots using the Wasserstein distance \citep{Wasserstein}. The Wasserstein distance is a distance function between distributions that can account for the distance between elements of two distributions, better known in recent years as the loss function of Wasserstein GAN (Generative Adversarial Network) \citep{WGAN}. Unlike Kullback-Leibler divergence, which is also a well-known measure of dissimilarity of probability distributions, it can be used even when the supports of the two discrete probability distributions do not completely overlap. In this study, for clustering based on the tendency of shots between players, but because the shots themselves are discrete, we employed this distance function, which can calculate the dissimilarity of discrete distributions. The following shows how to calculate the Wasserstein distance of two discrete probability distributions.

For two random variables $X, Y \in \mathbb{R}^{d}$, given a probability mass function with $m$ supports: $\mu(x)$ for $X = \{x_1, \ldots, x_m\}$ and a probability mass function with $n$ supports: $\nu(y)$ for $Y = \{y_1, \ldots, y_n\}$, and distance matrix $D \in \mathbb{R}^{m \times n}$ whose elements are the distances between $x_i$ and $y_j$ with parameter $p$ $(\geq1)$: $||x_i-y_j||_2^{p}$, the $p$-Wasserstein distance $W_p$ between $\mu$ and $\nu$ is defined as follows:

\begin{align}
\label{Wasserstein}
    W_p(\mu, \nu):= ({\operatorname{min}} & \sum_{i=1}^m \sum_{j=1}^n D_{ij} P_{ij})^{1/p}, 
\end{align}
where $P \in \mathbb{R}^{m \times n}$ is an asymmetric matrix representing the transportation plan, and $P_{ij}$ is the quantity (probability mass) to be transported from $x_i$ to $y_j$. In other words, equation (\ref{Wasserstein}) represents the total transportation cost in the most efficient case of carrying the probability mass of one distribution to match that of the other. In particular, when $p=1$, it is also called Earth Mover's Distance (EMD). In this study, Earth Mover's Distance was calculated for simplicity by considering each shot set for each player as an empirical discrete uniform distribution. In other words, we calculated the dissimilarity of the distributions $\mu$ and $\nu$ of the shots of the two players when $d=9$ (using principal components of shots) and $p=1$ (using distance matrix of Euclidean distance) in the above formula. Note that the data was limited to players for whom data for 30 or more shots existed, and 325 players were included in the analysis. A distance matrix of Earth Mover's Distances was created for the following clustering.

\begin{figure}[tb]
    \centering
    \includegraphics[scale=0.6]{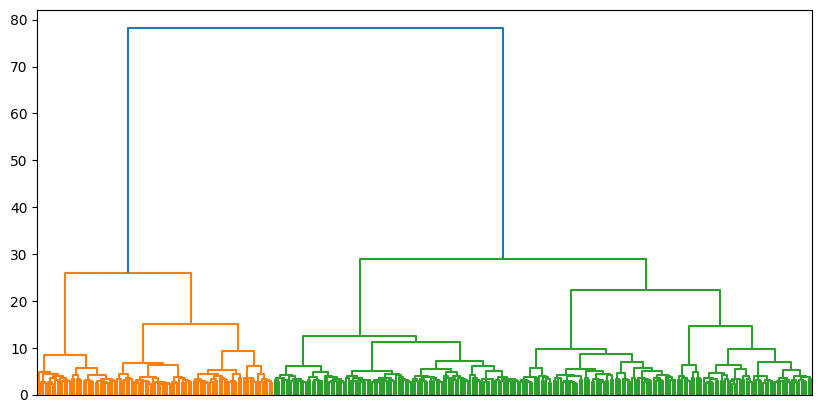}	
    \caption{Hierarchical clustering dendrogram of shooting style clustering. The vertical axis represents the 1-Wasserstein distance and the horizontal axis represents each player.}
    \label{fig:dendrogram}
\end{figure}

The players were then grouped using Ward's method \citep{Ward}, which is based on the calculated distance. Ward's method forms clusters so that the sum of squares of deviations within a cluster is minimized when merging two clusters. This method was used because it provides the most evenly distributed number of players in each cluster. From the dendrogram (Figure \ref{fig:dendrogram}), the best number of clusters was considered to be 2. To determine the appropriate number of clusters, silhouette coefficients were calculated. The silhouette coefficient $s^{(i)}$ is determined for data point $\boldsymbol{x}^{(i)}$ using the following degree of cohesion $a^{(i)}$ within the same cluster and the degree of deviation $b^{(i)}$ from the nearest separate cluster.

\begin{equation}
a^{(i)}=\frac{1}{\left|C_{\text {in}}\right|-1} \sum_{\boldsymbol{x}^{(j)} \in C_{\text {in}}} d(\boldsymbol{x}^{(i)}, \boldsymbol{x}^{(j)}),
\end{equation}

\begin{equation}
b^{(i)}=\frac{1}{|C_{\text {near}}|} \sum_{\boldsymbol{x}^{(j)} \in C_{\text {near}}} d(\boldsymbol{x}^{(i)}, \boldsymbol{x}^{(j)}),
\end{equation}

\begin{equation}
s^{(i)}=\frac{b^{(i)}-a^{(i)}}{\max(a^{(i)}, b^{(i)})},
\end{equation}
where $C_{\text {in}}$ is the cluster to which $\boldsymbol{x}^{(i)}$ belongs, $C_{\text {near}}$ is the nearest cluster from $\boldsymbol{x}^{(i)}$ other than $C_{\text {in}}$, and $d$ is an arbitrary distance function. The silhouette coefficient takes values in $[-1, 1]$ from the definition above and is closer to 1 the more agglomerated the data points are in the same cluster and the further apart the data points are in different clusters. In other words, the closer to 1 can be interpreted as better clustering. In this method, the information on the Earth Mover's Distance between data points (i.e., players) was used to calculate the mean silhouette coefficients when the number of clusters was between 2 and 20.

\begin{figure}[H]
    \centering
    \includegraphics[scale=0.6]{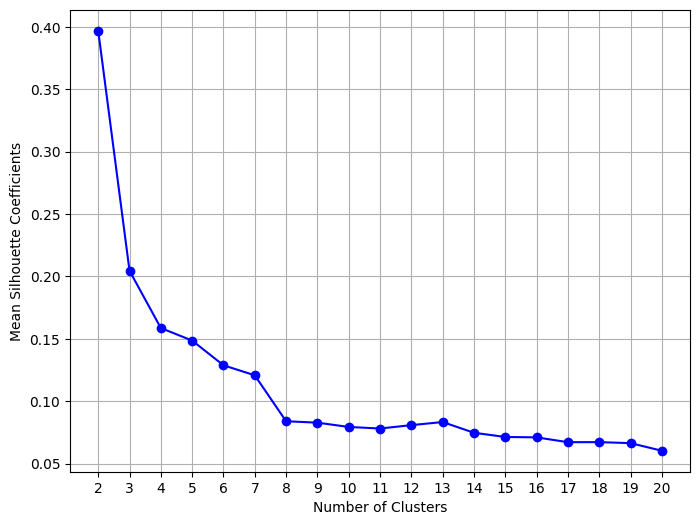}	
    \caption{Mean Silhouette Coefficients in Shooting Style Clustering }
    \label{fig:silhouette1}
\end{figure}

From the mean silhouette coefficients in Figure \ref{fig:silhouette1}, a small number of clusters was considered relatively good. However, since a small number of clusters does not provide meaningful information, 13 was adopted as the number of clusters just before the silhouette coefficient decreased, while also taking interpretability into account.

The attributes of each cluster were interpreted by looking at histograms of the features of the shots (before dimensionality reduction) contained in each cluster. This allowed us to confirm that the clustering results fit our intuition and to name the clusters.

As another verification, 
the shot efficiency of players between and within the clusters extracted was compared in True Shooting percentage (TS\%) for the 2015-16 season. TS\% is an advanced stat that represents the efficiency of the shot, calculated by the following formula: 

\begin{align}
    TS \%=\frac{PTS}{2 \times(FGA+0.44 \times FTA)} \times 100,
\end{align}
where PTS is total points scored, FGA is Field Goal Attempts, and FTA is Free Throw Attempts. Note that the possible range of TS\% is from 0 to 150. We examined the top five players in TS\% within each cluster and see if they fit our intuition. This analysis is limited to players with 200 or more FGA during the 2015-16 season to ensure the reliability of the TS\% value.

\subsubsection{Offensive Role Clustering} \label{sec:offensive_role_clustering}
\begin{figure}[H]
    \centering
    \includegraphics[scale=0.6]{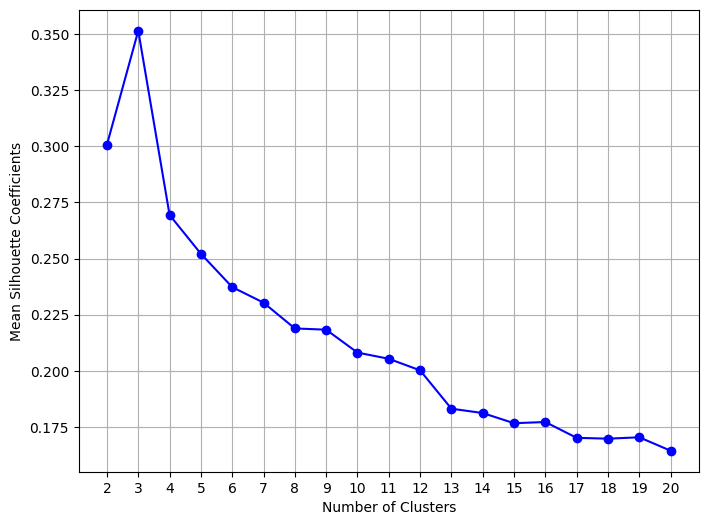}	
    \caption{Mean Silhouette Coefficients in Offensive Role Clustering \label{fig:silhouette2}}
\end{figure}

The K-means clustering algorithm was used first in this method. K-means is one of the most representative methods of non-hierarchical clustering. The silhouette coefficient was also used to calculate the number of clusters. (Figure \ref{fig:silhouette2}).

Again, a small number of clusters was considered desirable, but as mentioned earlier, that would be too small; thus, the number of clusters was set to 10, taking into account the interpretability and uniformity of the number of players in each cluster. However, it failed to consider complex offensive roles and included counter-intuitive results. Hierarchical clustering by Ward's method was tried as well as shooting styles, but these methods did not provide a resolution to the issue. Hence we employed Fuzzy C-means clustering \citep{c-means}, which allows for ambiguity (fuzziness) in membership to clusters. When applying Fuzzy C-means to a dataset consisting of $n$ data points $\boldsymbol{x}^{(i)}$, the following objective function is minimized.

\begin{figure}[H]
    \centering
    \includegraphics[scale=0.7]{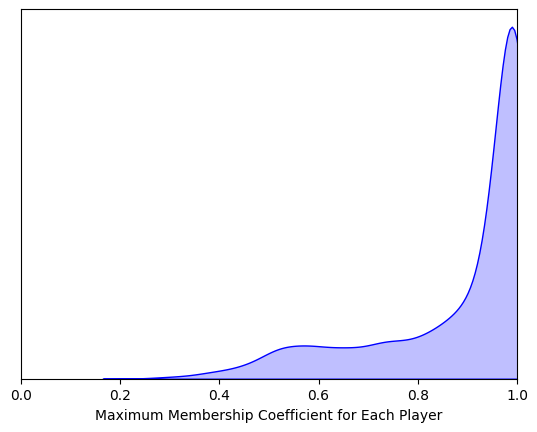}	
    \caption{Distribution of the Maximum Membership Coefficient for Each Player\label{fig:membership}}
\end{figure}

\begin{equation}
\label{fuzzy_cmeans_objective}
    J=\sum_{k=1}^c\sum_{i=1}^n\left(u_k^{(i)}\right)^q||\boldsymbol{x}^{(i)}-\boldsymbol{v}_k||^2,
\end{equation}
where $c$ is the number of clusters, $\boldsymbol{v}_k$ is the centroid of cluster $k$, and $u_k^{(i)}$ is the membership coefficient, in other words, the degree to which $\boldsymbol{x}_i$ belongs to cluster $k$. $l$ is a hyperparameter that controls fuzziness and takes a value greater than or equal to 1. In this objective function, the following equation is satisfied:

\begin{equation}
\label{membership_coefficient}
\sum_{k=1}^c u_k^{(i)} = 1.
\end{equation}
The higher the parameter $q$, the higher the fuzzy degree, i.e., $u_k^{(i)}$ for each cluster $k$ becomes more evenly distributed. In this case, Fuzzy C-means is applied with the objective of allowing data points that are far from any centroid in the K-means clustering results, i.e., players with composite offensive roles, to belong to multiple clusters. Therefore, the value of $q$ was selected to ensure that players close to the centroid were predominantly assigned to one cluster, while those positioned within the cluster's center were allowed to have membership in multiple clusters, with $q=1.2$ under $c=10$. Below is the distribution of the maximum membership coefficient for each player (Figure \ref{fig:membership}). Approximately 66\% of the players had a maximum membership coefficient of 0.9 or higher, and approximately 97\% had a maximum membership coefficient of 0.5 or higher.

The interpretation of the clusters extracted by the method here was done by the average value of each feature for players with a membership coefficient of 0.9 or higher. 
To validate the clustering results, the players with the low maximum membership coefficient (less than 0.6) were examined.  It is important to note that the validation approach differed from that of shooting style clustering because offensive role clustering is a soft clustering method and cannot be easily ranked.

\subsection{Variable Creation for Lineup Analysis} \label{sec:variable_creation}
In this study, we attempted to gain insight by interpreting a machine-learning model for the 5-player combination and a Bayesian model for the 2-player combination. The machine learning models have high predictive power but somewhat poor interpretability, so they do not tell us why those 5-player are better. Therefore, by creating a more interpretable statistical model, we quantitatively examined which combination was more effective (the effect of the 3-player combination was not examined in this study because the number of explanatory variables was too large for the data).

Hence, the methods used in the analysis of the 5-player playing style combination and the 2-player playing style combination are different, and therefore the explanatory variables used are also different. The former used the number of players of each cluster in the lineup and the latter used the number of combinations of clusters in the lineup as explanatory variables. For the former, it is simple: the number of explanatory variables is the number of clusters and the sum of each lineup is 5. The latter is more complex: the number of explanatory variables (the number of combination types) is the number of combinations when two clusters are taken out of all clusters, with repetition:

\begin{equation}
\label{number_of_explanatory_variables}
\left(\left(\begin{array}{l}
c \\
2
\end{array}\right)\right)=\left(\begin{array}{c}
c+1 \\
2
\end{array}\right) \text {, }
\end{equation}
where c is the number of clusters. The sum of the values of the explanatory variables for each lineup is the number of combinations of two players,

\begin{equation}
\left(\begin{array}{c}
5 \\
2
\end{array}\right)= 10\text {. }
\end{equation}
Note that the clustering method for shooting style was hard clustering and the clustering method for offensive roles was soft clustering, so the explanatory variables for the lineup data represented by shooting style were discrete values, and explanatory variables for the lineup represented by offensive roles have continuous values. In lineup analysis in terms of shooting style, because the amount of data was small relative to the number of clusters, some clusters were merged to reduce the number of parameters to be estimated (details are provided in Section \ref{sec:shooting_2player_combos}). In lineup analysis in terms of offensive role, each player belongs to a cluster softly, so the number $Combos_{kk^{\prime}}^{(i)}$ of combinations of clusters $k$ and $k^{\prime}$ ($k \leq k^{\prime})$ in a given lineup $i$ is calculated as follows:

\begin{equation}
\label{combos}    Combos_{kk^{\prime}}^{(i)}=\begin{cases} \sum_{a \in A^{(i)}}\sum_{a^{\prime} \in A^{(i)} s.t. a^{\prime} \neq a} u_k^{(a)} u_{k^{\prime}}^{(a^{\prime})} & \text{if }(k\neq k^{\prime}) \\
    \frac{1}{2}\sum_{a \in A^{(i)}}\sum_{a^{\prime} \in A^{(i)} s.t. a^{\prime} \neq a} u_k^{(a)} u_{k^{\prime}}^{(a^{\prime})} 
    & \text{if }(k = k^{\prime})
\end{cases}, 
\end{equation}
where $A^{(i)}$ is the set of players comprising lineup $i$ and $u_k^{(a)}$ is the membership coefficient of player $a$ for cluster $k$. Since we are considering combinations, we note the case $k=k^{\prime}$ separately. Actually, for each of the 10 combinations of players in the lineup, the column vector of membership coefficients for one player was multiplied by the row vector of membership coefficients for the other player to create a square matrix $M \in \mathbb{R}^{c \times c}$ ($c$: the number of clusters), whose diagonal elements $M_{kk}$ are the combinations of clusters $k$ and $M_{kk^{\prime}}$ and $M_{k^{\prime}k}$ are the combinations of clusters $k$ and $k^{\prime}$, and then added together as and the sum of the 10 combinations was used as the number of cluster combinations in the lineup. With the sum of the membership coefficients being 1 for each player, the total number of combinations after calculation in the lineup is still 10.

In this study, the lineup's offensive rating (OFFRTG) was used as the objective variable because our focus was solely on offense. Offensive rating is a stat that represents the expected points per 100 possessions, in other words, scoring efficiency. However, offensive ratings for lineups with short simultaneous playing time contain considerable noise due to the small sample size of the possessions. Conversely, limiting the analysis to lineups with long simultaneous playing time may introduce a bottleneck in lineup diversity, as frequently used lineups tend to indicate strength. To strike a balance, this study was limited to lineups that played more than 50 minutes of simultaneous time, or approximately 100 or more possessions, while lineups that played less than 300 minutes of simultaneous time used adjusted offensive rating. The adjustment method is similar to that used by \cite{kalman2020nba} and is as follows:

\begin{align}
    \textrm{adjustedOFFRTG} = \textrm{LineupOFFRTG}\times(\textrm{MIN}/300) + \textrm{TeamOFFRTG}\times(1-\textrm{MIN}/300),
\end{align}
where LineupOFFRTG is the offensive rating of the lineup, MIN is the minutes that the lineup was on the court at the same time, and TeamOFFRTG is the offensive rating of the team to which the lineup belongs. The number of lineup data used for the analysis with the results of the shooting style clustering was $n=995$, and the number of lineup data used for the analysis with the results of the offensive role clustering was $n=2270$.

\subsection{Lineup Analysis}
The main purpose of the analysis is to interpret regression models that predict scoring efficiency from the composition of lineups represented by clusters and to gain insight into which combinations of clusters are effective. The regression model here is a model that predicts an objective variable, OFFRTG, from explanatory variables that have information about how the lineup is structured. As mentioned in Section \ref{sec:variable_creation}, different methods are used in the analysis of a 5-player combination and a 2-player combination.

\subsubsection{Analysis of the combination of 5 players}
The method uses machine learning models with high predictive power to examine which lineups, represented by clusters, tend to be more effective in scoring.
First, we compared the predictive performance of OFFRTG with that of the regression model built with the lineup data represented by the clustering results of \cite{kalman2020nba}. Given the purpose of analyzing cluster compatibility, it is not necessary for the predictive performance of the regression model, based on lineup data represented by the clustering method used in this study, to surpass that of the previous study. While the interpretability of the clusters should offer new insights, excessively low prediction performance is still undesirable. This is because interpreting a model with significantly poor predictive performance for scoring efficiency does not necessarily mean that effective cluster combinations for increasing scoring efficiency are being examined. Therefore, as one criterion for measuring prediction performance, we applied the clustering method of \cite{kalman2020nba} to the target lineups described in Section \ref{sec:preprocessing}. Subsequently, we compared the prediction performance of the regression model constructed using lineup data represented by the results of the clustering in this study with that of the regression model developed using lineup data based on the clustering results of \cite{kalman2020nba}.

The regression models used were Support Vector Machine \citep{SVM}, LightGBM \citep{LightGBM}, and NGBoost \citep{NGBoost}. Support Vector Machine is a well-known machine learning model with high generalization capability. LightGBM is a gradient-boosting decision tree well known for its lightweight and high predictive performance that is often used in data analysis competitions. NGBoost, like LightGBM, is a gradient-boosting model (in this study, decision trees were used as the base model), but with the advantage of easily computing prediction uncertainty. Root mean squared error (RMSE), mean absolute error (MAE), and negative log likelihood (NLL; only NGBoost is capable of outputting predictive distributions) were used as evaluation metrics. Of the lineup data, one team was used as validation data, the remaining 29 teams were used as training data for cross-validation. The evaluation metrics for prediction against the validation data were averaged. The hyperparameters were manually adjusted to determine the settings that yielded the best average evaluation metrics across all 30 teams.
Second, we sorted the lineups represented by the clustering results in this study in the order of their predicted values using NGBoost to observe trends. Teams and lineups with large NLL values, i.e., large prediction errors, were also examined (NGBoost was chosen because it allows for the incorporation of prediction uncertainty, which was considered important for this analysis).

\subsubsection{Analysis of the combination of 2 players}
In this method, the effect of the combination of two players is estimated quantitatively. Specifically, the partial regression coefficients of the explanatory variable, the combination of clusters created in the preprocessing, are estimated for each team. In mathematical terms, for each team $t (=1, \ldots, 30)$, the posterior distributions of the fixed effects $\alpha_{t}$ (this can be seen as an effect of the average team style) and vectors of combination effects of two clusters $\boldsymbol{\beta}_{t}$ of the teams were estimated by MCMC (Markov chain Monte Carlo method) with NUTS (No-U-turn Sampler; \citeauthor{NUTS}, \citeyear{NUTS}) based on the following Bayesian hierarchical linear regression model \citep{bayesian_hierarchical_modeling}:

\begin{equation}
\label{hierarchical_bayes}
\begin{gathered}
y^{(i)} \sim \mathcal{N}\left(\mu^{(i)}, {\epsilon^{(i)}}^2\right), \\
\mu^{(i)}=\alpha_t+{\boldsymbol{x}_{t}^{(i)}}^T \boldsymbol{\beta}_t, \\
\alpha_t \sim \mathcal{N}\left(\mu_\alpha,10^2\right), \\
\beta_t \sim \mathcal{N}\left(\mu_\beta, \sigma_\beta^2\right), \\
\mu_\beta \sim \mathcal{N}\left(0,10^2\right), \\
\sigma_\beta \sim \operatorname{Half-Normal}(10), \\
\epsilon^{(i)} \sim \operatorname{Half-Normal}(10),
\end{gathered}
\end{equation}
where $y^{(i)}$ is the (adjusted) OFFRTG of lineup $i$ and $\boldsymbol{x}_t^{(i)}$ is the vector of explanatory variables for lineup $i$ belonging to a team $t$ (its elements are the number of cluster combinations in the lineup as noted in Section \ref{sec:variable_creation}. Its value is discrete greater than or equal to 0 for the analysis in terms of shooting style and $Combos_{kk^{\prime}}^{(i)}$ in equation \ref{combos} for the analysis in terms of offensive role). $\mu_\alpha$, the mean of the prior distribution of $\alpha_t$, was set with reference to the (adjusted) OFFRTG distribution, for the analysis in terms of shooting style, $\mu_\alpha=105$, and for the analysis in terms of offensive role, $\mu_\alpha=110$. As for the convergence of MCMC, we judged by $\hat{R}$ \citep{rhat}.
We took the expected value of the estimated posterior distribution and calculated the median of that value among all teams to compare the effects of the combinations. By estimating the effect of combinations on a team-by-team basis and taking the median value, it is robust to a particular combination of players. Note that the super-prior distribution was not set for the parameters of the prior distribution of $\alpha_t$ because MCMC was not stable.

\section{Results} \label{sec:results}
\subsection{Shooting Style Clustering}
\subsubsection{Cluster Interpretation}

\begin{figure}[H]
  \begin{minipage}[b]{0.48\linewidth}
    \centering
    \includegraphics[keepaspectratio, scale=0.5]{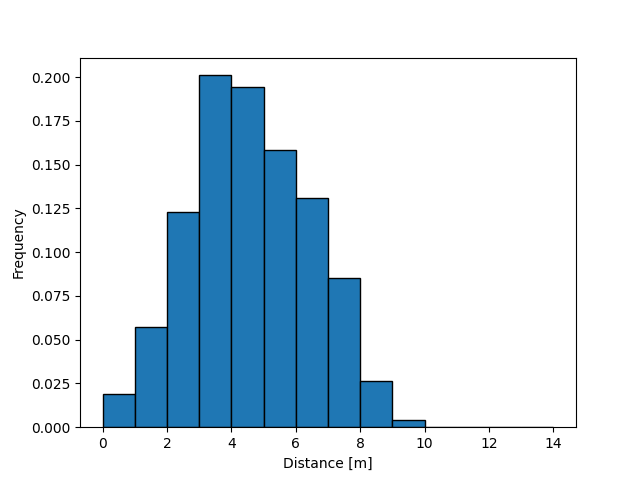}
  \end{minipage}
  \begin{minipage}[b]{0.48\linewidth}
    \centering
    \includegraphics[keepaspectratio, scale=0.5]{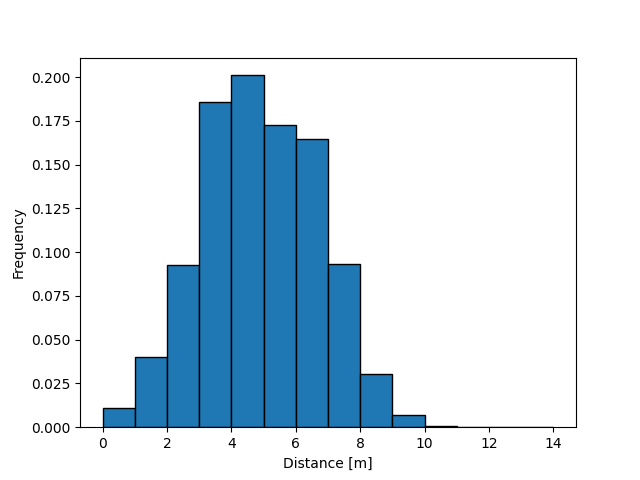}
  \end{minipage}
  \caption{Distance between the shooter and the rim 3 seconds before shot in two clusters as examples. Normalized frequencies are displayed. On the left is Cluster 1, which tends to have the shortest distance from the rim 3 seconds before the shot. On the right is Cluster 2, which also tends to have a shorter distance from the rim 3 seconds before the shot.
  }
  \label{fig:initialdistance}
\end{figure}

\begin{figure}[H]
  \begin{minipage}[b]{0.48\linewidth}
    \centering
    \includegraphics[keepaspectratio, scale=0.5]{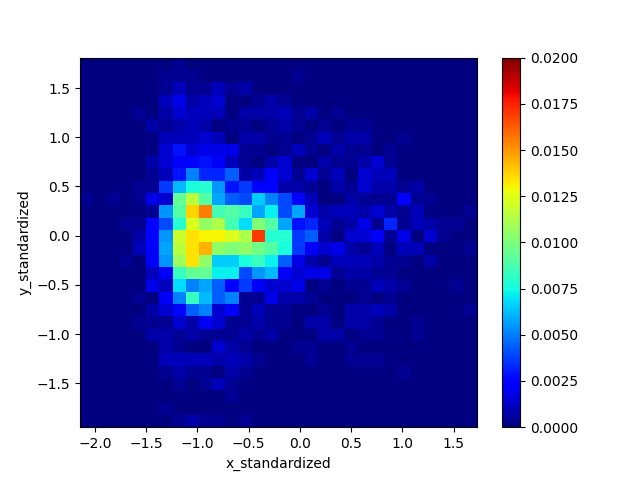}
  \end{minipage}
  \begin{minipage}[b]{0.48\linewidth}
    \centering
    \includegraphics[keepaspectratio, scale=0.5]{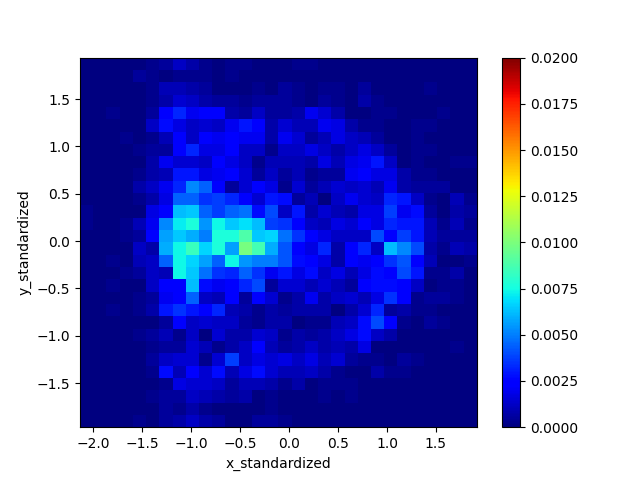}
  \end{minipage}
  \caption{2D histogram of shot location, in two clusters as examples. Normalized frequencies are displayed. On the left is Cluster 1, with shot locations concentrated near the rim. On the right is Cluster 2, which exhibits shots well distributed from the rim to mid-range.}
  \label{fig:ShotLocation}
\end{figure}

In shooting style clustering, the individual clusters were interpreted using the features of the shots in each cluster. For example, regarding the distance between the shooter and the rim 3 seconds before the shot in clusters 1 and 2 (Figure \ref{fig:initialdistance}), the players in these clusters were often about 4 meters from the rim in common. Compared to the other clusters, the players in these clusters are more likely to be near the rim before the shot, indicating that the player in these clusters is a Big Man. Regarding the location of the shot (Figure \ref{fig:ShotLocation}), both clusters prefer to shoot near the rim, but compared to cluster 1, players in cluster 2 tend to shoot mid-range shots as well. From the above, we named cluster 1 ``Close-range Big'' and cluster 2 ``Mid-range Big''. For all clusters, we also confirmed that the clusters to which the players belonged were not counterintuitive. 

\begin{table}[H]
\caption{Shooting Style Cluster Description}
\centering
\label{tab:shootingstyleclusterdescription}
\scalebox{0.8}{
\begin{tabular}{p{0.5cm}cp{7cm}cc} 
\hline
No. & Cluster Name                                                         & Description                                                                                              & Average Height {[}cm{]} & Example Players                                                                                       \\ \hline 1 &
\begin{tabular}[c]{@{}c@{}}\\Close-range Big\\ (CB)\end{tabular}       & Big Men who attempt most shots from close-range.\newline               & 210.6                   & \begin{tabular}[c]{@{}c@{}}Andre Drummond \\ Dwight Howard \end{tabular}                \\ \hline 2 &
\begin{tabular}[c]{@{}c@{}}\\Mid-range Big\\ (MB)\end{tabular}         & Big Men who can shoot from close-range as well as mid-range. \newline& 209.9                   & \begin{tabular}[c]{@{}c@{}}Pau Gasol\\ LaMarcus Aldridge\end{tabular}                   \\ \hline 3 &
\begin{tabular}[c]{@{}c@{}}\\Mid-range All-rounder\\ (MA)\end{tabular} & Players who play offense from mid-range and often shoot from near the high post.       \newline                  & 208.9                   & \begin{tabular}[c]{@{}c@{}} Kevin Garnet \\ Dirk Nowitzki\end{tabular}               \\ \hline 4 &
\begin{tabular}[c]{@{}c@{}}\\Mid-range Slasher\\ (MS)\end{tabular}     & Players who attack the rim from mid-range through post-play or drive.         \newline                           & 206.5                   & \begin{tabular}[c]{@{}c@{}} Giannis Antetokounmpo \\ DeMarcus Cousins\end{tabular} \\ \hline 5 &
\begin{tabular}[c]{@{}c@{}}\\Outside All-rounder\\ (OA)\end{tabular}   & Players who begin offense from the outside and aim to shoot from anywhere with their versatile skills. & 198.6                   & \begin{tabular}[c]{@{}c@{}}Stephen Curry \\ Kevin Durant\end{tabular}                  \\ \hline 6 & \begin{tabular}[c]{@{}c@{}}\\Pull-up Ball-Handler\\ (PH)\end{tabular}  & Players who aim for many pull-up jumpers.                                                                & 189.9                   & \begin{tabular}[c]{@{}c@{}}Kyle Lowry \\ Damian Lillard\end{tabular}                  \\ \hline 7 & \begin{tabular}[c]{@{}c@{}}\\Drive Ball-Handler\\ (DH)\end{tabular}    & Players who often drive from the outside, but also attempt 3-pointers moderately.    \newline                     & 190.7                   & \begin{tabular}[c]{@{}c@{}}Russell Westbrook \\ James Harden \\  \end{tabular}             \\ \hline 8 &
\begin{tabular}[c]{@{}c@{}}\\Stretch Four\\ (S4)\end{tabular}          & Big shooter who often attempts corner threes or threes from the top position.     \newline                       & 205.2                   & \begin{tabular}[c]{@{}c@{}} Nikola Mirotic \\ Meyers Leonard\end{tabular}             \\ \hline 9 &
\begin{tabular}[c]{@{}c@{}}\\Corner Shooter\\ (CS)\end{tabular}        & Shooter attempting mainly corner threes.                                                                    & 198.6                   & \begin{tabular}[c]{@{}c@{}} Patrick Beverley \\ Jason Terry\end{tabular}                \\ \hline 10 &
\begin{tabular}[c]{@{}c@{}}\\Pure Shooter\\ (PS)\end{tabular}          & Shooter who attempts threes from any location.                                                               & 194.4                   & \begin{tabular}[c]{@{}c@{}}Eric Gordon \\ Kyle Korver \end{tabular}                       \\ \hline 11 & \begin{tabular}[c]{@{}c@{}}\\Outside Slasher\\ (OS)\end{tabular}       & Players who do not shoot many threes and prefer to drive from the outside.      \newline                         & 199.0                   & \begin{tabular}[c]{@{}c@{}}DeMar DeRozan \\ Tony Parker\end{tabular}                   \\ \hline 12 &
\begin{tabular}[c]{@{}c@{}}\\Drive Shooter\\ (DS)\end{tabular}         & Similar to Corner Shooter, but shooter with a slight preference for drive.      \newline                         & 201.5                   & \begin{tabular}[c]{@{}c@{}} Klay Thompson \\ Vince Carter\end{tabular}                  \\ \hline 13 & \begin{tabular}[c]{@{}c@{}}\\Stretch All-rounder\\ (SA)\end{tabular}   & Players who shoot from close to mid-range, stretch and shoot threes as well.      \newline                       & 205.7                   & \begin{tabular}[c]{@{}c@{}} Kristaps Porzingis \\ Kevin Love\end{tabular}                \\ \hline

\end{tabular}
}
\end{table}

\begin{figure}[H]
    \centering
    \includegraphics[width=0.8\linewidth]{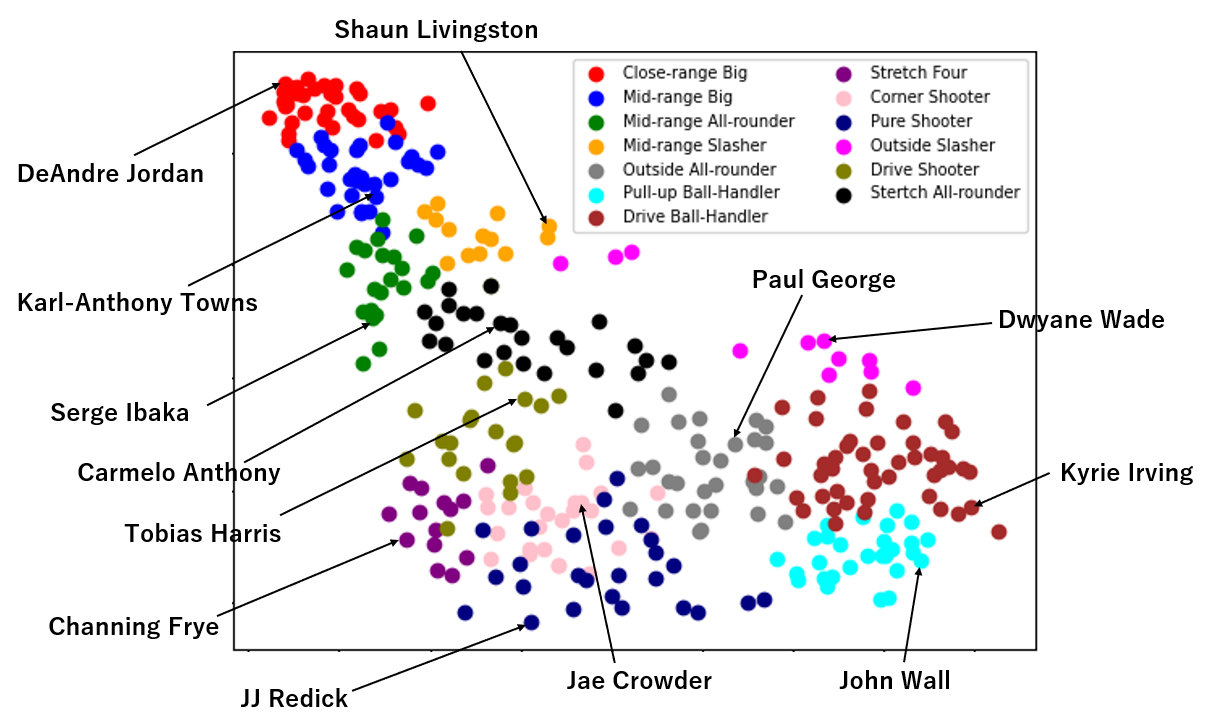}	
    \caption{Visualization of shooting style clusters by t-SNE}
    \label{fig:clusters}
\end{figure}

Table \ref{tab:shootingstyleclusterdescription} describes the name of each cluster, the players' average height, examples of players, and a brief description. The third cluster was named ``Mid-range All-rounder'' due to their tendency to attempt mid-shots, especially near the high post. The fourth cluster was named ``Mid-range Slasher'' owing to the tendency to hold the ball slightly longer than clusters 1 and 2, although it prefers shots near the rim. The fifth cluster was named ``Outside All-rounder'' because it tended to aim for shots from any location and was interpreted as being aggressive in driving while attempting triples. The sixth cluster was named ``Pull-up Ball-Hander'' due to the tendency to hold the ball longer and has more threes from outside the corners and shots just inside its arc. The seventh cluster was named ``Drive Ball-Handler'' because, like cluster 6, it had a longer ball-holding time and more shots were made near the rim. The eighth cluster was named the ``Stretch Four'' due to its higher frequency of 3-pointers from the corners and top and a slightly taller average height. The ninth cluster was named ``Corner Shooter because of its preference for shooting threes, especially from the corner. The tenth cluster preferred the 3-point, but unlike the ninth cluster, it tended to aim from any position, hence the name ``Pure Shooter''. The eleventh cluster was named ``Outside Slasher'' because players tended to hold the ball longer and preferred to shoot near the rim but not so much for 3-pointers. The twelfth cluster was similar to the ninth cluster but was named ``Drive Shooter'' due to a slight preference for shots near the rim. The 13th cluster prefers close-range to mid-range shots, but can also attempt a reasonable number of 3-pointers, thus we named it ``Stretch All-rounder''.
Based on the distance matrix, the players were plotted in a 2-dimensional coordinate plane with cluster labels using t-SNE (Figure \ref{fig:clusters}). In this figure, players with similar shot tendencies are placed closer together, while players with dissimilar shot tendencies are placed farther apart.

\subsubsection{Comparison of Shot Efficiency within Cluster}
Tables \ref{tab:top5ofTSforCB}, \ref{tab:top5ofTSforOAandSA}, and \ref{tab:top5ofTSforOSandDS} show the top five TS\% players and their respective field goal attempts for several clusters in the 2015-16 season (the higher the value, the more reliable the value of the TS\%). The remaining clusters are listed in the Supplementary Materials. From the tables, we can see that many of the players are from teams that were strong at the time, such as Golden State Warriors (GSW), San Antonio Spurs (SAS), and Oklahoma City Thunder (OKC). This suggests that players from stronger teams tended to have better shooting efficiency (when viewed in conjunction with remaining clusters in the Supplementary Materials, many players from other strong teams of the time, the Los Angeles Clippers (LAC) and the Toronto Raptors (TOR), are also represented), which fits our intuition.

\begin{table}[H]
\caption{Top 5 players for Close-range Big in TS\%.}
\label{tab:top5ofTSforCB}
\centering
\begin{tabular}{ll}
\begin{tabular}{rrr}
\hline
\multicolumn{3}{c}{Close-range Big}                                                                          \\ \hline
\multicolumn{1}{c}{Player Name (Team)}             & \multicolumn{1}{l}{FGA} & \multicolumn{1}{l}{TS\%} \\ \hline
\multicolumn{1}{r}{Hassan Whiteside (MIA)} & \multicolumn{1}{r}{682} & 62.9                               \\
\multicolumn{1}{r}{DeAndre Jordan (LAC)}   & \multicolumn{1}{r}{508} & 62.8                               \\
\multicolumn{1}{r}{Cole Aldrich (LAC)}     & \multicolumn{1}{r}{225} & 62.6                               \\
\multicolumn{1}{r}{Andrew Bogut (GSW)}     & \multicolumn{1}{r}{279} & 62.3                               \\
\multicolumn{1}{r}{Steven Adams (OKC)}     & \multicolumn{1}{r}{426} & 62.1                               \\ \hline
\end{tabular}
\end{tabular}
\end{table}

\begin{table}[H]
\caption{Top 5 players for Outside All-rounder and Stretch All-rounder in TS\%.}
\centering
\label{tab:top5ofTSforOAandSA}
\begin{tabular}{ll}
\begin{tabular}{rrr}
\hline
\multicolumn{3}{c}{Outside All-rounder}                                                                     \\ \hline
\multicolumn{1}{c}{Player Name (Team)}             & \multicolumn{1}{l}{FGA}  & \multicolumn{1}{l}{TS\%} \\ \hline
\multicolumn{1}{l}{Stephen Curry (GSW)}    & \multicolumn{1}{r}{1598} & 66.9                               \\
\multicolumn{1}{l}{Kevin Durant (OKC)}     & \multicolumn{1}{r}{1381} & 63.4                               \\
\multicolumn{1}{l}{Chandler Parsons (DAL)} & \multicolumn{1}{r}{651}  & 58.9                               \\
\multicolumn{1}{l}{Omri Casspi (SAC)}      & \multicolumn{1}{r}{622}  & 58.7                               \\
\multicolumn{1}{l}{Evan Fournier (ORL)}    & \multicolumn{1}{r}{929}  & 58.7                               \\ \hline
\end{tabular}

&

\begin{tabular}{lrr}
\hline
\multicolumn{3}{c}{Stretch All-rounder}                                                                  \\ \hline
\multicolumn{1}{c}{Player Name (Team)}           & \multicolumn{1}{l}{FGA}  & \multicolumn{1}{l}{TS\%} \\ \hline
\multicolumn{1}{l}{Kawhi Leonard (SAS)}   & \multicolumn{1}{r}{1090}  & 61.6                               \\
\multicolumn{1}{l}{Draymond Green (GSW)}     & \multicolumn{1}{r}{819} & 58.7                               \\
\multicolumn{1}{l}{Mike Scott (ATL)} & \multicolumn{1}{r}{376}  & 57.5                               \\
\multicolumn{1}{l}{Chris Bosh (MIA)} & \multicolumn{1}{r}{767} & 57.1                               \\
\multicolumn{1}{l}{Kelly Olynyk (BOS)}   & \multicolumn{1}{r}{556} & 56.1                               \\ \hline

\end{tabular}
\end{tabular}
\end{table}

\begin{table}[H]
\caption{Top 5 players for Outside Slasher and Drive Shooter in TS\%.}
\centering
\label{tab:top5ofTSforOSandDS}
\begin{tabular}{ll}
\begin{tabular}{rrr}
\hline
\multicolumn{3}{c}{Outside Slasher}                                                                          \\ \hline
\multicolumn{1}{c}{Player Name (Team)}              & \multicolumn{1}{l}{FGA} & \multicolumn{1}{l}{TS\%} \\ \hline
\multicolumn{1}{l}{Jonathon Simmons (SAS)} & \multicolumn{1}{r}{242} & 58.6                               \\
\multicolumn{1}{l}{DeMar DeRozan (TOR)}        & \multicolumn{1}{r}{1377} & 55                                \\
\multicolumn{1}{l}{Tony Parker (SAS)}       & \multicolumn{1}{r}{710} & 54.6                               \\
\multicolumn{1}{l}{Donald Sloan (BKN)}      & \multicolumn{1}{r}{350} & 53.6                               \\
\multicolumn{1}{l}{James Johnson (TOR)}  & \multicolumn{1}{r}{240} & 53.2                               \\ \hline
\end{tabular}

&

\begin{tabular}{lrr}
\hline
\multicolumn{3}{c}{Drive Shooter}                                                                          \\ \hline
\multicolumn{1}{c}{Player Name (Team)}            & \multicolumn{1}{l}{FGA} & \multicolumn{1}{l}{TS\%} \\ \hline
\multicolumn{1}{l}{Klay Thompson (GSW)}    & \multicolumn{1}{r}{1386} & 59.7                              \\
\multicolumn{1}{l}{Thabo Sefolosha (ATL)} & \multicolumn{1}{r}{372} & 57.8                               \\
\multicolumn{1}{l}{Alonzo Gee (NOP)} & \multicolumn{1}{r}{255} & 57.2                               \\
\multicolumn{1}{l}{Andre Roberson (OKC)} & \multicolumn{1}{r}{777} & 57.1                               \\
\multicolumn{1}{l}{Andre Iguodala (GSW)}  & \multicolumn{1}{r}{609} & 56.4                               \\ \hline

\end{tabular}
\end{tabular}
\end{table}

\subsection{Linup Analysis in terms of Shooting Style}
\subsubsection{Analysis of the combination of 5 players}
First, we used the clustering method of \cite{kalman2020nba} using various 23 stats, and the results described above. We created lineup data expressed as the number of players in each cluster and compared the predictive performance of adjusted OFFRTG using three different regression models. Although we could not replicate all nine clusters of \cite{kalman2020nba}, we could replicate six, which we used to create a regression model. Tables \ref{tab:metrics_of_shootingstyle_lineup} and \ref{tab:metrics_of_kalman_lineup1} show the respective evaluation metrics. Compared to the results using the methods of \cite{kalman2020nba}, the results using the shooting style clusters were less accurate in predicting OFFRTG. This may reflect differences in whether or not information on players' scoring ability was included when clustering However, these results do not dissuade us from pursuing further analysis.

\begin{table}[H]
\centering
\caption{Evaluation metrics for regression models with lineup represented by 13 clusters of shooting styles}
\label{tab:metrics_of_shootingstyle_lineup}
\begin{tabular}{cccc}                                                                         \hline
\multicolumn{1}{c}{Model}                            & RMSE           & MAE            & NLL \\ \hline
\multicolumn{1}{c}{Support Vector Machine}         & 4.516 & 3.706 &  -   \\
\multicolumn{1}{c}{LightGBM}                       & 4.594         & 3.786         & -              \\
\multicolumn{1}{c}{NGBoost}                       & 4.565          & 3.759          & 2.978                \\ \hline
\end{tabular}
\end{table}

\begin{table}[htbp]
\centering
\caption{Evaluation metrics for regression models with lineups (of same as the shooting style) represented by 9 clusters extracted by the method of \cite{kalman2020nba} using 23 various stats}
\label{tab:metrics_of_kalman_lineup1}
\begin{tabular}{cccc}
                               \hline
\multicolumn{1}{c}{Model}                            & RMSE           & MAE            & NLL \\ \hline
\multicolumn{1}{c}{Support Vector Machine}         & 4.380 & 3.575 &  -   \\
\multicolumn{1}{c}{LightGBM}                       & 4.443         & 3.636         & -              \\
\multicolumn{1}{c}{NGBoost}                       & 4.335         & 3.537          & 2.918                \\ \hline
\end{tabular}
\end{table}

\begin{figure}[h]
    \centering
    \includegraphics[width=0.63\linewidth]{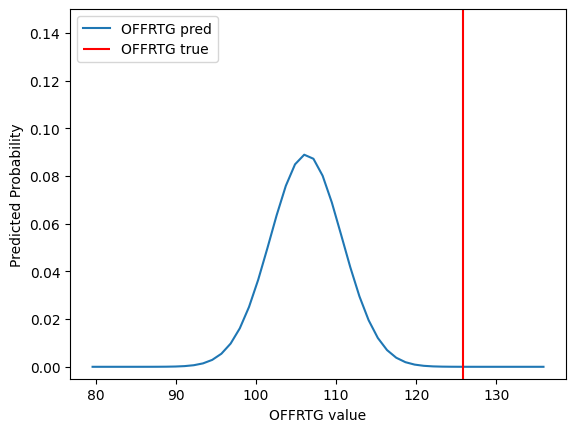}	
    \caption{Predicted distribution and the true value of lineup represented by shooting style cluster with the largest NLL; lineup consisting of A. Iguodala, S. Curry, K. Thompson, H. Barnes, and D. Green in GSW for the 2015-16 season}
    \label{fig:pred_dist1}
\end{figure}

Next, we sorted lineups in descending or ascending order based on the predicted OFFRTG values to see what trends were observed. Tables \ref{tab:top10lineups1} and \ref{tab:bottom10lineups1} show the top 10 and bottom 10 predicted OFFRTG lineups, including actual player names. Top 10 lineups show that in addition to one Big Man, there is typically one ball-handler and one shooter with the other two being all-rounders. From this result, it appears advantageous to incorporate a wider variety of shooting styles or to introduce more variations in the shots taken by players. The bottom 10 lineup was that of SAS, which tended to underestimate OFFRTG. SAS was a strong team with a unique team style at the time. We also looked at the team with the largest NLL average, and that team was GSW, another strong team at the time. Figure \ref{fig:pred_dist1} shows the predicted distribution and true value of OFFRTG for the lineups with the largest NLL. GSW also tended to run a particular team offense with respect to this lineup. From the aforementioned findings, predicting the scoring efficiency of lineups from teams with a general tendency not to run a team offense proves challenging with shooting style clusters. This poor predictability could also be attributed to the high shooting ability of each player.

\begin{table}[H]
\centering
\caption{Lineups of shooting style clusters with the top 10 OFFRTG predicted values}
\label{tab:top10lineups1}
\scalebox{0.85}{
\begin{tabular}{llllllll}
\hline
Rank & Season  & Team & Player 1    & Player 2    & Player 3     & Player 4         & Player 5         \\ \hline
1    & 2012-13 & DEN  & A. Iguodala & C. Brewer   & D. Gallinari & T. Lawson        & K. Faried   \\ 
2    & 2013-14 & NYK  & T. Chandler & C. Anthony  & R. Felton    & A. Bargnani      & I. Shumpert      \\ 
3    & 2013-14 & NYK  & T. Chandler & C. Anthony  & J. Smith     & R. Felton        & I. Shumpert      \\ 
4    & 2016-17 & DET  & I. Smith    & T. Harris   & J. Leuer     & A. Drummond      & K. Caldwell-Pope \\ 
5    & 2015-16 & DET  & M. Morris   & T. Harris   & R. Jackson   & A. Baynes        & K. Caldwell-Pope \\ 
6    & 2015-16 & DET  & M. Morris   & R. Jackson  & A. Drummond  & K. Caldwell-Pope & S. Johnson       \\ 
7    & 2016-17 & DET  & M. Morris   & T. Harris   & R. Jackson   & A. Drummond      & S. Johnson       \\ 
8    & 2016-17 & DET  & B. Udrih    & M. Morris   & J. Leuer     & A. Baynes        & S. Johnson       \\ 
9    & 2015-16 & DET  & E. İlyasova & M. Morris   & R. Jackson   & A. Drummond      & S. Johnson       \\ 
10    & 2015-16 & DET  & S. Blake    & A. Tolliver & M. Morris    & A. Baynes        & S. Johnson      \\ \hline
\end{tabular}
}
\end{table}

\begin{table}[H]
\scalebox{0.85}{
\begin{tabular}{lllllllllllllll}
\hline
CB & MB & MA & MS & OA & PH & DH & S4 & PS & CS & OS & DS & SA & OFFRTG\_true & OFFRTG\_predicted \\ \hline
1  & 0  & 0  & 0  & 1  & 0  & 1  & 0  & 1  & 0  & 0  & 1  & 0  & 109.120      & 107.344           \\
1  & 0  & 1  & 0  & 0  & 1  & 0  & 0  & 1  & 0  & 0  & 0  & 1  & 104.877      & 107.343           \\
1  & 0  & 0  & 0  & 1  & 1  & 0  & 0  & 1  & 0  & 0  & 0  & 1  & 108.846      & 107.343           \\
1  & 0  & 0  & 0  & 1  & 0  & 1  & 0  & 0  & 0  & 0  & 1  & 1  & 104.310      & 107.342           \\
1  & 0  & 0  & 0  & 1  & 0  & 1  & 0  & 0  & 0  & 0  & 1  & 1  & 103.063      & 107.342           \\
1  & 0  & 0  & 0  & 2  & 0  & 1  & 0  & 0  & 0  & 0  & 0  & 1  & 106.873      & 107.342           \\
1  & 0  & 0  & 0  & 1  & 0  & 1  & 0  & 0  & 0  & 0  & 1  & 1  & 104.988      & 107.342           \\
1  & 0  & 0  & 0  & 1  & 0  & 1  & 0  & 0  & 0  & 0  & 0  & 2  & 102.791      & 107.342           \\
1  & 0  & 0  & 0  & 1  & 0  & 1  & 1  & 0  & 0  & 0  & 0  & 1  & 104.756      & 107.342           \\
1  & 0  & 0  & 0  & 1  & 0  & 0  & 1  & 0  & 1  & 0  & 0  & 1  & 105.193      & 107.342           \\ \hline
\end{tabular}
}
\end{table}

\begin{table}[H]
\centering
\caption{Lineups of shooting style clusters with the bottom 10 OFFRTG predicted values}
\label{tab:bottom10lineups1}
\scalebox{0.85}{
\begin{tabular}{llllllll}

\hline
Rank & Season  & Team & Player 1    & Player 2     & Player 3     & Player 4      & Player 5    \\ \hline
995  & 2018-19 & SAS  & L. Aldridge & R. Gay       & M. Belinelli & D. DeRozan    & P. Mills    \\ 
994  & 2017-18 & SAS  & P. Gasol    & L. Aldridge  & D. Green     & P. Mills      & K. Anderson \\ 
993  & 2016-17 & SAS  & P. Gasol    & L. Aldridge  & D. Green     & P. Mills      & K. Leonard  \\ 
992  & 2018-19 & SAS  & L. Aldridge & M. Belinelli & D. DeRozan   & D. Cunningham & P. Mills    \\ 
991  & 2016-17 & SAS  & D. Lee      & L. Aldridge  & D. Green     & P. Mills      & K. Leonard  \\ 
990  & 2013-14 & SAS  & M. Ginobili & B. Diaw      & M. Belinelli & T. Splitter   & P. Mills    \\ 
989  & 2015-16 & SAS  & M. Ginobili & D. West      & B. Diaw      & D. Green      & P. Mills    \\ 
988  & 2013-14 & SAS  & M. Ginobili & B. Diaw      & M. Belinelli & P. Mills      & A. Baynes   \\ 
987  & 2014-15 & SAS  & M. Ginobili & B. Diaw      & M. Belinelli & T. Splitter   & P. Mills    \\ 
986  & 2014-15 & SAS  & M. Ginobili & B. Diaw      & M. Belinelli & P. Mills      & A. Baynes   \\ \hline
\end{tabular}
}
\end{table}

\begin{table}[H]
\centering
\scalebox{0.85}{
\begin{tabular}{lllllllllllllll}

\hline
CB & MB & MA & MS & OA & PH & DH & S4 & PS & CS & OS & DS & SA & OFFRTG\_true & OFFRTG\_predicted \\ \hline
0  & 1  & 0  & 0  & 0  & 0  & 0  & 0  & 0  & 2  & 1  & 0  & 1  & 109.005      & 105.140           \\
0  & 2  & 0  & 0  & 0  & 0  & 0  & 0  & 0  & 2  & 1  & 0  & 0  & 107.464      & 105.140           \\
0  & 2  & 0  & 0  & 0  & 0  & 0  & 0  & 0  & 2  & 0  & 0  & 1  & 111.017      & 105.140           \\
0  & 1  & 0  & 0  & 0  & 0  & 0  & 0  & 0  & 2  & 1  & 1  & 0  & 111.357      & 105.140           \\
1  & 1  & 0  & 0  & 0  & 0  & 0  & 0  & 0  & 2  & 0  & 0  & 1  & 108.733      & 105.140           \\
1  & 1  & 0  & 0  & 1  & 0  & 0  & 0  & 0  & 2  & 0  & 0  & 0  & 110.800      & 105.325           \\
0  & 2  & 0  & 0  & 1  & 0  & 0  & 0  & 0  & 2  & 0  & 0  & 0  & 106.469      & 105.325           \\
1  & 1  & 0  & 0  & 1  & 0  & 0  & 0  & 0  & 2  & 0  & 0  & 0  & 113.568      & 105.325           \\
1  & 1  & 0  & 0  & 1  & 0  & 0  & 0  & 0  & 2  & 0  & 0  & 0  & 104.875      & 105.325           \\
1  & 1  & 0  & 0  & 1  & 0  & 0  & 0  & 0  & 2  & 0  & 0  & 0  & 106.002      & 105.325           \\ 
\hline
\end{tabular}
}
\end{table}

\subsubsection{Analysis of the combination of 2 players} \label{sec:shooting_2player_combos}
In this analysis, we first merged some of the clusters to the extent that interpretability is not lost (to reduce the number of parameters to be estimated and to stabilize MCMC). Specifically, the Mid-range Big, Mid-range All-rounder, and Mid-range Slasher located in the upper left of Figure \ref{fig:clusters} are integrated into the ``Mid-range'', the Stretch All-rounder and Outside All-rounder located in the middle of Figure \ref{fig:clusters} are integrated into the ``All-rounder'', and the Pull-up Ball-Handler, Drive Ball-Handler, and Outside Slasher located in the lower right of Figure \ref{fig:clusters} are integrated into the ``Ball-handler'', and the Pure Shooter, Corner Shooter, and Drive Shooter, which are located at the bottom left of Figure \ref{fig:clusters}, are integrated to form ``3point-shooter''. Only Close-range Big was not integrated, only renamed as ``Close-range''. Hence, the number of clusters is 5, and from equation \ref{number_of_explanatory_variables}, the number of explanatory variables is 15. The posterior distribution of the parameters was estimated by MCMC based on equation \ref{hierarchical_bayes}. Since $\hat{R} < 1.1$ for all parameters, MCMC was considered to have converged. Table \ref{tab:betas1} shows the medians among all teams for means of the posterior distribution of the partial regression coefficient $\beta_t$ for each combination, and some combinations tend to produce positive effects and others less so. Detailed discussions are given in Section \ref{sec:discussion}.

\begin{table}[H]
\caption{Estimated effects of combinations of shooting styles}
\label{tab:betas1}
\centering
\begin{tabular}{ll}

\hline
Combination type                     & Median of E[$\beta_t$] \\ \hline
All-rounder \& 3point-shooter    & 0.1715               \\ \hline
All-rounder \& Ball-handler      & 0.1715               \\ \hline
Ball-handler \& 3point-shooter   & 0.1610                \\ \hline
Ball-handler \& Ball-handler     & 0.1080                \\ \hline
Mid-range \& All-rounder         & 0.1070                \\ \hline
3point-shooter \& 3point-shooter & 0.1000                  \\ \hline
Close-range \& All-rounder       & 0.0935               \\ \hline
All-rounder \& All-rounder       & 0.0920                \\ \hline
Close-range \& Close-range       & 0.0915               \\ \hline
Mid-range \& Ball-handler        & 0.0875               \\ \hline
Close-range \& Mid-range         & 0.0840                \\ \hline
Mid-range \& Mid-range           & 0.0710                \\ \hline
Close-range \& Ball-handler      & 0.0385               \\ \hline
Mid-range \& 3point-shooter      & 0.0345               \\ \hline
Close-range \& 3point-shooter    & -0.0055              \\ \hline
\end{tabular}
\end{table}

\subsection{Offensive Role Clustering}
\subsubsection{Cluster Interpretation}
In the offensive role clustering, the clusters were interpreted and named based on the average of the features of the players with a membership coefficient of at least 90\% for each cluster. We also confirmed that the players in each cluster were intuitive. Table \ref{tab:offensiveroleclusterdescription} lists the name of each cluster, its relatively high stats, a brief description, and examples of players whose membership coefficient was 90\% or higher in the 2022-23 season.

\begin{table}[H]
\small
\centering
\caption{Offensive Role Cluster Description}
\label{tab:offensiveroleclusterdescription}
\scalebox{0.88}{
\begin{tabular}{clll}
\hline
Cluster Name                                                            & High Stats                                                                                      & Description                                                                                                                                                                                                                                                  & Example Player                                                            \\ \hline
\begin{tabular}[c]{@{}c@{}}Stretch Big \\ (STB)\end{tabular}            & \begin{tabular}[c]{@{}l@{}}Spot-up\%: 27\% \\ PnR roll-man\%:  19\% \end{tabular}       & \begin{tabular}[c]{@{}l@{}}Big Men who play the role of screener \\ but prefers to catch-and-shoot. \end{tabular}                                                                                                            & \begin{tabular}[c]{@{}l@{}}Brook Lopez\\ Jaren Jackson Jr.\end{tabular}   \\ \hline
\begin{tabular}[c]{@{}c@{}}Isolation Attacker\\ (ISA)\end{tabular}      & \begin{tabular}[c]{@{}l@{}}Isolation\%: 14\% \\ USG\%: 28\% \\ \end{tabular}                       & \begin{tabular}[c]{@{}l@{}}Players who prefer to play isolation offense \\ and play a central role in the team.\\ They tend to have a variety of offensive skills.\end{tabular}                                                                               & \begin{tabular}[c]{@{}l@{}}LeBron James\\ Kawhi Leonard\end{tabular}      \\ \hline
\begin{tabular}[c]{@{}c@{}}Post-up Big \\ (PUB)\end{tabular}            & \begin{tabular}[c]{@{}l@{}}Post-up\%: 26\%\\ PnR roll-man\%: 18\%\end{tabular}     & \begin{tabular}[c]{@{}l@{}}Players who prefer post play.\\ This number is decreasing \\ every year.\end{tabular}                                                                                                                                         & \begin{tabular}[c]{@{}l@{}}Anthony Davis\\ James Wiseman\end{tabular}     \\ \hline
\begin{tabular}[c]{@{}c@{}}Secondary Ball-Handler \\ (SBH)\end{tabular} & \begin{tabular}[c]{@{}l@{}}PnR ball-handler\%: 36\%\\ Spot-up\%: 23\%\end{tabular}   & \begin{tabular}[c]{@{}l@{}}Ball-handlers with slightly more Spot-up.\\ Compared to PBH, PnR ball-handler\% \\ is about 10\% lower and Spot-up\% \\ is about 10\% higher.\end{tabular} & \begin{tabular}[c]{@{}l@{}}Malcolm Brogdon\\ Dennis Schroder\end{tabular} \\ \hline
\begin{tabular}[c]{@{}c@{}}Transition Attacker\\ (TRA)\end{tabular}     & \begin{tabular}[c]{@{}l@{}}Transition\%: 23\%\\ Spot-up\%: 33\%\end{tabular}                    & \begin{tabular}[c]{@{}l@{}}Players who like to play offense in transition.\\ They also often play offense from Spot-up.\end{tabular}                                                                                                                      & \begin{tabular}[c]{@{}l@{}}Caleb Martin\\ Alex Caruso\end{tabular}        \\ \hline
\begin{tabular}[c]{@{}c@{}}Primary Ball-Handler\\ (PBH)\end{tabular}    & \begin{tabular}[c]{@{}l@{}}PnR ball-handler\%: 46\%\\ AST\%: 33\%\end{tabular}       & \begin{tabular}[c]{@{}l@{}}Ball-handlers who often use on-ball screens. \\ They are the primary offensive initiator.\end{tabular}                                                                                                                            & \begin{tabular}[c]{@{}l@{}}LaMelo Ball\\ Tyrese Haliburton\end{tabular}   \\ \hline
\begin{tabular}[c]{@{}c@{}}Spot-up Shooter\\ (SUS)\end{tabular}         & Spot-up\%: 47\%                                                                               & \begin{tabular}[c]{@{}l@{}}Shooters who prefer to shoot from Spot-up. \\ This number is increasing every year.\end{tabular}                                                                                                                           & \begin{tabular}[c]{@{}l@{}}Yuta Watanabe\\ P.J. Tucker\end{tabular}       \\ \hline
\begin{tabular}[c]{@{}c@{}}Roll \& Cut Big\\ (RCB)\end{tabular}         & \begin{tabular}[c]{@{}l@{}}Cut\%: 26\%\\ PnR roll-man\%: 23\%\end{tabular}         & \begin{tabular}[c]{@{}l@{}}Players who dive to the rim to score points. \\ They are also good at Putback.\end{tabular}                                                                                                                                         & \begin{tabular}[c]{@{}l@{}}Clint Capela\\ Rudy Gobert\end{tabular}        \\ \hline
\begin{tabular}[c]{@{}c@{}}Off-screen Shooter\\ (OSS)\end{tabular}      & \begin{tabular}[c]{@{}l@{}}Hand-off\%+Off-screen\%: \\34\%\\ Spot-up\%: 28\%\end{tabular}     & \begin{tabular}[c]{@{}l@{}}Players who use Hand-off and Off-screen \\ to get open shots.\\ Compared to SUS, there are fewer players.\end{tabular}                                                                                               & \begin{tabular}[c]{@{}l@{}}Kevin Huerter\\ Duncan Robinson\end{tabular}   \\ \hline
\begin{tabular}[c]{@{}c@{}}Wing with Handle\\ (WWH)\end{tabular}        & \begin{tabular}[c]{@{}l@{}}PnR ball-handler\%: 22\%\\ Spot-up\%: 29\%\end{tabular} & \begin{tabular}[c]{@{}l@{}}Players who excel at drive and 3-pointers. \\ Their playing style is somewhere \\ between a ball-handler and a shooter.\end{tabular}                                                                               & \begin{tabular}[c]{@{}l@{}}Kyle Kuzma\\ Seth Curry\end{tabular}           \\ \hline
\end{tabular}
}
\end{table}

\subsubsection{Low Maximum Membership Coefficient Players}
Before the analysis of offensive efficiency, we highlight several players whose membership coefficients were at a maximum value of 0.6 or less in the 2022-23 season. First, Karl-Anthony Towns of the Minnesota Timberwolves exhibited a membership coefficient of 0.58 for Stretch Big, 0.12 for Post-up Big, 0.11 for Wing with Handle, among others. This potentially reflects his playing style as a Big Man with high shooting and ball-handling skills. Second, Patty Mills of the Brooklyn Nets exhibited a membership coefficient of 0.55 for Wing with Handle and 0.45 for Off-screen Shooter. This is a good reflection of his style of shooting with high accuracy while utilizing his fast movement (in the previous season, he had an off-screen shooter of 0.98; This is likely related to the role of his other teammates). Third, Lauri Markkanen of the Utah Jazz, who won the NBA Most Improved Player Award for the 2022-23 season, had a membership coefficient of 0.4 for Off-screen Shooter, 0.26 for Transition Attacker, and 0.24 for Stretch Big, among others. This may reflect his playing style, characterized by less dribbling and more off-ball movement, as well as a preference for a variety of offenses. Traditional clustering methods like Ward's method of hierarchical clustering and K-means clustering were inadequate to reflect the complex offensive roles listed above, prompting the use of soft clustering methods such as Fuzzy C-means. It's worth noting that while Fuzzy C-means was applied to shooting style clustering as well, interpretable clusters could not be obtained.

\subsection{Linup Analysis in terms of Offensive Role}
\subsubsection{Analysis of the combination of 5 players}
We initially compared the evaluation metrics of the regression model predicting adjusted OFFRTG for lineups using the clustering results of \cite{kalman2020nba} with those of the regression model predicting adjusted OFFRTG for lineups using the offensive role. Tables \ref{tab:metrics_of_offensiverole_lineup} and \ref{tab:metrics_of_kalman_lineup2} present the respective outcomes indicating that although offensive clustering does not include stats on scoring, it is more accurate in predicting scoring efficiency compared to the clusters of \cite{kalman2020nba}.

\begin{table}[H]
\centering
\caption{Evaluation metrics for regression models with lineups represented by offensive role clusters}
\label{tab:metrics_of_offensiverole_lineup}
\begin{tabular}{cccc}                                                                         \hline
\multicolumn{1}{c}{Model}                            & RMSE           & MAE            & NLL \\ \hline
\multicolumn{1}{c}{Support Vector Machine}         & 4.825 & 3.932 &  -   \\
\multicolumn{1}{c}{LightGBM}                       & 4.854         & 3.955         & -              \\
\multicolumn{1}{c}{NGBoost}                       & 4.786          & 3.869          & 3.066                \\ \hline
\end{tabular}
\end{table}

\begin{table}[H]
\centering
\caption{Evaluation metrics for regression models with lineups (of the same as the offensive role)  represented by 9 clusters extracted by the method of \cite{kalman2020nba} using 23 various stats}
\label{tab:metrics_of_kalman_lineup2}
\begin{tabular}{cccc}
                               \hline
\multicolumn{1}{c}{Model}                            & RMSE           & MAE            & NLL \\ \hline
\multicolumn{1}{c}{Support Vector Machine}         & 4.921 & 4.027 &  -   \\
\multicolumn{1}{c}{LightGBM}                       & 4.866         & 3.959         & -              \\
\multicolumn{1}{c}{NGBoost}                       & 4.850         & 3.942          & 3.089                \\ \hline
\end{tabular}
\end{table}

\begin{figure}[H]
    \centering
    \includegraphics[width=0.63\linewidth]{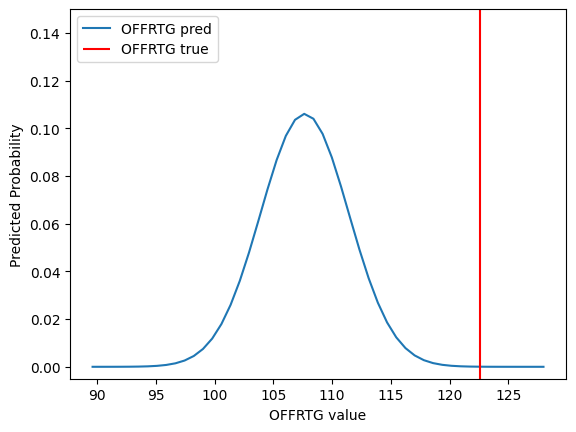}	
    \caption{Predicted distribution and the true value of lineup represented by offensive role cluster with the largest NLL; lineup consisting of P. Millsap, W. Barton, N. Jokic, J. Murray, and M. Porter Jr. in DEN for the 2020-21 season}
    \label{fig:pred_dist2}
\end{figure}

Next, we sorted lineups in descending or ascending order based on the predicted OFFRTG values to see what trends were observed. Tables \ref{tab:top10lineups2} and \ref{tab:bottom10lineups2} show the top 10 and bottom 10 predicted OFFRTG lineups, including actual player names. Table \ref{tab:top10lineups2} shows that lineups with high predicted values tend to consist mainly of about two Isolation Attackers, one or two Spot-up Shooters and Off-screen Shooters combined. Conversely, Table \ref{tab:bottom10lineups2} shows that lineups with lower predicted values tend to consist mainly of one or two Stretch Big and Post-up Bigs combined and one Primary Ball-Handler.
We also looked at the team with the least successful prediction of OFFRTG for the lineups represented in the offensive role cluster and found that it was DEN. The predicted distribution and true value of the lineup with the largest NLL in DEN are shown in Figure \ref{fig:pred_dist2}. The most likely cause is that the offensive role clusters are insufficient to reflect Nikola Jokic's playing style. This suggests that his playing style is idiosyncratic in that he prefers to Post-up but also likes to assist his teammates with passes.

\begin{table}[H]
\centering
\caption{Lineups of offensive role clusters with the top 10 OFFRTG predicted values}
\label{tab:top10lineups2}
\scalebox{0.85}{
\begin{tabular}{llllllll}
\hline
Rank & Season  & Team & Player 1    & Player 2  & Player 3    & Player 4    & Player 5         \\  \hline
1    & 2022-23 & BOS  & A. Horford  & J. Brown  & J. Tatum    & D. White    & G. Williams     \\
2    & 2020-21 & LAC  & S. Ibaka    & N. Batum  & P. Beverley & P. George   & K. Leonard      \\
3    & 2022-23 & BKN  & K. Durant   & K. Irving & J. Harris   & R. O'Neale  & N. Claxton      \\
4    & 2022-23 & LAC  & P. George   & M. Morris & K. Leonard  & I. Zubac    & T. Mann         \\
5    & 2020-21 & LAC  & P. Beverley & P. George & M. Morris   & K. Leonard  & I. Zubac        \\
6    & 2021-22 & BKN  & K. Durant   & K. Irving & A. Drummond & S. Curry    & B. Brown        \\
7    & 2022-23 & BOS  & A. Horford  & M. Smart  & J. Brown    & J. Tatum    & G. Williams     \\
8    & 2021-22 & MIA  & P. Tucker   & J. Butler & D. Dedmon   & D. Robinson & T. Herro        \\
9    & 2018-19 & GSW  & A. Iguodala & K. Durant & S. Curry    & K. Thompson & D. Green        \\
10   & 2022-23 & BOS  & A. Horford  & J. Brown  & J. Tatum    & D. White    & R. Williams III \\ \hline
\end{tabular}
}
\end{table}

\begin{table}[H]
\scalebox{0.85}{
\begin{tabular}{llllllllllll}
\hline
STB  & ISA  & PUB  & SBH  & TRA  & PBH  & SUS  & RCB  & OSS  & WWH  & OFFRTG\_true & OFFRTG\_pred \\ \hline
0.00 & 1.57 & 0.00 & 0.02 & 0.00 & 0.00 & 2.00 & 0.00 & 0.00 & 1.40 & 116.325      & 116.783      \\
0.80 & 1.99 & 0.15 & 0.00 & 0.02 & 0.00 & 1.96 & 0.04 & 0.00 & 0.01 & 118.108      & 116.371      \\
0.02 & 1.73 & 0.02 & 0.03 & 0.02 & 0.01 & 1.98 & 1.00 & 0.06 & 0.14 & 110.116      & 116.087      \\
0.00 & 1.99 & 0.00 & 0.00 & 1.03 & 0.00 & 0.97 & 1.00 & 0.00 & 0.00 & 122.993      & 115.924      \\
0.05 & 1.99 & 0.00 & 0.00 & 0.24 & 0.00 & 1.68 & 1.00 & 0.00 & 0.03 & 117.988      & 115.823      \\
0.30 & 2.00 & 0.01 & 0.02 & 0.67 & 0.00 & 0.01 & 1.00 & 0.01 & 0.98 & 115.765      & 115.779      \\
0.00 & 1.58 & 0.00 & 0.08 & 0.03 & 0.00 & 2.00 & 0.00 & 0.00 & 1.31 & 113.422      & 115.731      \\
0.31 & 1.93 & 0.00 & 0.06 & 0.09 & 0.01 & 0.59 & 1.00 & 1.00 & 0.00 & 115.007      & 115.650      \\
0.29 & 1.96 & 0.01 & 0.02 & 1.59 & 0.00 & 0.05 & 0.00 & 1.00 & 0.07 & 121.111      & 115.597      \\
0.00 & 1.57 & 0.00 & 0.02 & 0.01 & 0.00 & 1.00 & 1.00 & 0.00 & 1.40 & 118.197      & 115.514      \\ \hline
\end{tabular}
}
\end{table}

\begin{table}[H]
\centering
\caption{Lineups of offensive role clusters with the bottom 10 OFFRTG predicted values}
\label{tab:bottom10lineups2}
\scalebox{0.85}{
\begin{tabular}{llllllll}
\hline
Rank & Season  & Team & Player 1     & Player 2     & Player 3            & Player 4            & Player 5     \\ \hline
2270 & 2018-19 & BKN  & J. Harris    & D. Russell   & R. Hollis-Jefferson & J. Allen            & R. Kurucs   \\
2269 & 2015-16 & DEN  & D. Arthur    & G. Harris    & J. Sampson          & N. Jokic            & E. Mudiay   \\
2268 & 2018-19 & BKN  & A. Crabbe    & J. Harris    & D. Russell          & R. Hollis-Jefferson & J. Allen    \\
2267 & 2016-17 & ATL  & D. Howard    & T. Sefolosha & P. Millsap          & K. Bazemore         & D. Schroder \\
2266 & 2016-17 & ATL  & K. Korver    & D. Howard    & P. Millsap          & K. Bazemore         & D. Schroder \\
2265 & 2015-16 & CHI  & P. Gasol     & M. Dunleavy  & D. Rose             & T. Gibson           & E. Moore    \\
2264 & 2016-17 & MIN  & R. Rubio     & G. Dieng     & A. Wiggins          & K. Towns            & K. Dunn     \\
2263 & 2018-19 & NOP  & J. Holiday   & E. Moore     & A. Davis            & E. Payton           & J. Randle   \\
2262 & 2015-16 & DEN  & K. Faried    & G. Harris    & J. Sampson          & N. Jokic            & E. Mudiay   \\
2261 & 2016-17 & PHI  & S. Rodriguez & G. Henderson & R. Covington        & J. Embiid           & J. Okafor  \\ \hline
\end{tabular}
}
\end{table}

\begin{table}[H]
\scalebox{0.85}{
\begin{tabular}{llllllllllll}
\hline
STB  & ISA  & PUB  & SBH  & TRA  & PBH  & SUS  & RCB  & OSS  & WWH  & OFFRTG\_true & OFFRTG\_pred \\ \hline
1.00 & 0.00 & 0.00 & 0.00 & 1.00 & 1.00 & 0.00 & 1.00 & 1.00 & 0.00 & 110.900      & 103.042      \\
1.00 & 0.00 & 1.00 & 0.00 & 1.08 & 1.00 & 0.00 & 0.00 & 0.88 & 0.04 & 101.120      & 103.386      \\
1.00 & 0.00 & 0.00 & 0.00 & 0.00 & 1.00 & 0.00 & 1.00 & 2.00 & 0.00 & 108.172      & 103.949      \\
0.07 & 0.00 & 1.78 & 0.00 & 1.00 & 1.00 & 0.00 & 0.15 & 0.00 & 1.00 & 97.400       & 104.034      \\
0.07 & 0.00 & 1.78 & 0.00 & 0.00 & 1.00 & 0.00 & 0.15 & 1.00 & 1.00 & 98.320       & 104.477      \\
0.02 & 0.00 & 1.42 & 0.00 & 0.00 & 1.00 & 0.00 & 0.56 & 1.00 & 1.00 & 108.095      & 104.517      \\
1.00 & 0.88 & 1.00 & 1.72 & 0.00 & 0.37 & 0.00 & 0.00 & 0.00 & 0.03 & 109.925      & 104.524      \\
0.18 & 0.12 & 1.77 & 0.89 & 0.04 & 1.00 & 0.00 & 0.00 & 0.97 & 0.04 & 117.273      & 104.555      \\
0.00 & 0.00 & 1.00 & 0.00 & 1.08 & 1.00 & 0.00 & 1.00 & 0.88 & 0.04 & 111.012      & 104.563      \\
0.00 & 0.00 & 1.99 & 0.02 & 1.00 & 0.98 & 0.00 & 0.00 & 0.04 & 0.95 & 100.283      & 104.819      \\ \hline
\end{tabular}
}
\end{table}

\subsubsection{Analysis of the combination of 2 players}
Based on equation \ref{hierarchical_bayes}, the posterior distribution of the parameters was estimated by MCMC (the number of explanatory variables is 55 from equation \ref{number_of_explanatory_variables}). Since $\hat{R}<1.1$ for all parameters, MCMC was deemed to have converged. Tables \ref{tab:top15effects} and \ref{tab:bottom15effects} show the top 15 and bottom 15 estimated effects of the combinations. The results suggest that combinations that include Isolation Attacker, a strong scorer, and Spot-up Shooter, a player who often goes for 3-pointers from Spot-up, tend to produce positive effects while combinations that include Stretch Big, which favors pop-out, tend to produce negative effects. Detailed discussions are given in Section \ref{sec:discussion}.

\begin{table}[H]
\centering
\caption{Top 15 estimated effects of combinations of offensive roles}
\label{tab:top15effects}
\begin{tabular}{ll}
\hline
Combination type & Median of E[$\beta_t$]   \\ \hline
ISA \& WWH  & 0.3460 \\ \hline
ISA \& TRA  & 0.2900 \\ \hline
PBH \& SUS  & 0.2055 \\ \hline
SUS \& RCB  & 0.1710 \\ \hline
WWH \& WWH  & 0.1400 \\ \hline
PUB \& SUS  & 0.1315 \\ \hline
SUS \& WWH  & 0.1310 \\ \hline
ISA \& PBH  & 0.1255 \\ \hline
SBH \& PBH  & 0.0960 \\ \hline
ISA \& PUB  & 0.0925 \\ \hline
TRA \& SUS  & 0.0910 \\ \hline
ISA \& SUS  & 0.0900 \\ \hline
PBH \& RCB  & 0.0890 \\ \hline
STB \& ISA  & 0.0560 \\ \hline
RCB \& WWH  & 0.0535 \\ \hline
\end{tabular}
\end{table}

\begin{table}[H]
\centering
\caption{Bottom 15 estimated effects of combinations of offensive roles}
\label{tab:bottom15effects}
\begin{tabular}{ll}
\hline
Combination type &  Median of E[$\beta_t$]   \\ \hline
PUB \& WWH  & -0.4720 \\ \hline
STB \& TRA  & -0.4145 \\ \hline
STB \& PUB  & -0.3710 \\ \hline
SBH \& RCB  & -0.3170 \\ \hline
STB \& WWH  & -0.2890 \\ \hline
STB \& PBH  & -0.2780 \\ \hline
PUB \& PBH  & -0.2715 \\ \hline
RCB \& OSS  & -0.2380 \\ \hline
STB \& SBH  & -0.2005 \\ \hline
TRA \& RCB  & -0.1995 \\ \hline
STB \& OSS  & -0.1880 \\ \hline
STB \& SUS  & -0.1755 \\ \hline
TRA \& OSS  & -0.1710 \\ \hline
TRA \& WWH  & -0.1565 \\ \hline
SBH \& TRA  & -0.1515 \\ \hline
\end{tabular}
\end{table}

\section{Discussion} \label{sec:discussion}
In this study, we defined the players' shooting styles and offensive roles by clustering and used the results to analyze the compatibility between the clusters. The clustering results for shooting style were deemed more precise in defining shooting style, as 10 more clusters were extracted compared to the results of \cite{Fan2023}, and each cluster matched well with intuitive expectations.
The results of the analysis of the effects (compatibility) of the cluster combinations (Tables \ref{tab:top10lineups1}, \ref{tab:bottom10lineups1} and \ref{tab:betas1}) showed that the combination of All-rounder, who shoots a variety of shots, Ball-handler, who shoots from drive, and 3point-shooter, who mainly shoots threes, is effective. Conversely, combinations of Bigs, such as mid-range players who mainly aim mid-range shots, and combinations of mid-range and close-range players who mainly aim close-range shots, tend not to yield positive effects. However, as shown in the SAS lineups in Table \ref{tab:bottom10lineups1}, even lineups that include combinations that are considered less effective will not reduce scoring efficiency much if the players have high shot ability.

The results of the offensive role clustering were considered to be able to reflect the composite playing style of the players based on the qualitative evaluation.
In the analysis of the effects of offensive role combinations, Tables \ref{tab:top10lineups2}, and \ref{tab:top15effects} show that the combination of Isolation Attacker, strong scorers who are the core of the offense, with wings, ball-handlers, and shooters who support them, is effective. While previous research \citep{kalman2020nba} indicated the effectiveness of having one strong scorer, our findings suggest that having two strong scorers can further enhance offensive performance. We also found that Spot-up Shooter tend to produce positive effects as well. This suggests that they are effectively expanding their space.
From Tables \ref{tab:bottom10lineups2} and \ref{tab:bottom15effects},  We also found that the combination of Post-up Big and Stretch Big, Bigs who like Cutting relatively little, with wings, ball-handlers, and shooters tends to be less effective. This may be due to their inability to effectively shake up the opposing defense.

The proposed clustering offers valuable applications in various contexts. Shooting styles are useful for understanding the playing styles of unfamiliar players because they can easily capture the shooting tendencies of specific players. Offensive styles are similarly useful, but also for guiding lineup formation (based on comparisons with previous study \citep{kalman2020nba}).

\section{Conclusion}
In this study, we introduced new player clustering methods for analyzing offensive lineups: shooting style clustering using tracking data, and offensive role clustering based on annotated playtypes and statistics. For the former, we used interpretable hand-crafted shot features and Wasserstein distances between players' shot distributions. 
For the latter, we applied soft clustering to annotated playtype data for the first time. 
By analyzing the combination of five effective players based on machine learning model predictions and the effect of two-player combinations based on Bayesian hierarchical models, we unveiled new insights into the composition of lineups and their efficiency in scoring.

This study was limited to providing general findings of player compatibility in an offense. Practically, if it were possible to estimate them simultaneously while separating the compatibility of two specific players from the general playstyle compatibility, it would help determine whether a player is replaceable or not for a given player. In addition, the clusters extracted do not necessarily best represent the current characteristics of the player, especially for shooting style clustering, because the data were limited to the 2015-16 season. If you want to capture a player's most recent shooting style, clustering should be done with much more recent data.

	\bibliography{refs}

\newpage

\newif\ifarxiv
\arxivfalse 

\ifarxiv
\renewcommand{\thesection}{\Alph{section}}
\setcounter{section}{0}
\setcounter{figure}{0}
\setcounter{table}{0}
\else



\fi

\newpage
\section*{}
\vspace{20mm}
\Large{\bf{Supplementary materials \\
\\

\noindent } }

\vspace{10mm}
\ifarxiv
\noindent\large{Kazuhiro Yamada and Keisuke Fujii}
\else
\noindent\large{}
\fi

\newpage
\section{Supplementary Tables}
The following table shows the top 5 players with TS\% of clusters that could not be included due to space limitations.

\begin{table}[h]
\centering
\caption{Top 5 players for Mid-range Big and Pull-up Ball-Handler in TS\%.}
\begin{tabular}{ll}
\begin{tabular}{rrr}
\hline
\multicolumn{3}{c}{Mid-range Big}                                                                      \\ \hline
\multicolumn{1}{c}{Player Name (Team)}         & \multicolumn{1}{l}{FGA} & \multicolumn{1}{l}{TS\%} \\ \hline
\multicolumn{1}{l}{Enes Freedom (OKC)} & \multicolumn{1}{r}{719} & 62.6                               \\
\multicolumn{1}{l}{Brandon Bass (BOS)} & \multicolumn{1}{r}{317} & 61.9                               \\
\multicolumn{1}{l}{Ian Mahinmi (IND)}  & \multicolumn{1}{r}{448} & 60.3                               \\
\multicolumn{1}{l}{Amir Johnson (BOS)} & \multicolumn{1}{r}{426} & 60.2                               \\
\multicolumn{1}{l}{Gorgui Dieng (MIN)} & \multicolumn{1}{r}{579} & 60.1                               \\ \hline

\end{tabular}

&

\begin{tabular}{lrr}
\hline
\multicolumn{3}{c}{Pull-up Ball-Handler}                                                                  \\ \hline
\multicolumn{1}{c}{Player Name (Team)}           & \multicolumn{1}{l}{FGA}  & \multicolumn{1}{l}{TS\%} \\ \hline
\multicolumn{1}{l}{Lou Williams (LAL)}   & \multicolumn{1}{r}{693}  & 58.4                               \\
\multicolumn{1}{l}{Kyle Lowry (TOR)}     & \multicolumn{1}{r}{1198} & 57.8                               \\
\multicolumn{1}{l}{Jerryd Bayless (MIL)} & \multicolumn{1}{r}{437}  & 56.8                               \\
\multicolumn{1}{l}{Damian Lillard (POR)} & \multicolumn{1}{r}{1474} & 56.0                               \\
\multicolumn{1}{l}{Kemba Walker (CHA)}   & \multicolumn{1}{r}{1331} & 55.4                               \\ \hline

\end{tabular}
\end{tabular}
\end{table}

\begin{table}[h]
\centering
\caption{Top 5 players for Mid-range All-rounder and Mid-range Slasher in TS\%.}
\begin{tabular}{ll}
\begin{tabular}{rrr}
\hline
\multicolumn{3}{c}{Mid-range All-rounder}                                                        \\ \hline
\multicolumn{1}{c}{Player Name (Team)}          & \multicolumn{1}{l}{FGA} & \multicolumn{1}{l}{TS\%} \\ \hline
\multicolumn{1}{l}{Al Horford (ATL)}    & \multicolumn{1}{r}{719} & 56.5                               \\
\multicolumn{1}{l}{Anthony Davis (NOP)} & \multicolumn{1}{r}{317} & 55.9                               \\
\multicolumn{1}{l}{Dwight Powell (DAL)} & \multicolumn{1}{r}{448} & 55.7                               \\
\multicolumn{1}{l}{Dirk Nowitzki (DAL)} & \multicolumn{1}{r}{426} & 55.5                               \\
\multicolumn{1}{l}{Blake Griffin (LAC)} & \multicolumn{1}{r}{579} & 54.4                               \\ \hline
\end{tabular}

&

\begin{tabular}{lrr}
\hline
\multicolumn{3}{c}{Mid-range Slasher}                                                                     \\ \hline
\multicolumn{1}{c}{Player Name (Team)}                  & \multicolumn{1}{l}{FGA}  & \multicolumn{1}{l}{{TS\%}} \\ \hline
\multicolumn{1}{l}{Shaun Livingston (GSW)}      & \multicolumn{1}{r}{379}  & 58.1                               \\
\multicolumn{1}{l}{Andrew Nicholson (ORL)}      & \multicolumn{1}{r}{310}  & 56.7                               \\
\multicolumn{1}{l}{Joffrey Lauvergne (DEN)}     & \multicolumn{1}{r}{380}  & 56.7                               \\
\multicolumn{1}{l}{Giannis Antetokounmpo (MIL)} & \multicolumn{1}{r}{1013} & 56.6                               \\
\multicolumn{1}{l}{Richaun Holmes (PHI)}        & \multicolumn{1}{r}{222}  & 56.4                               \\ \hline

\end{tabular}
\end{tabular}
\end{table}

\begin{table}[htbp]
\centering
\caption{Top 5 players for Drive Ball-Handler and Stretch Four in TS\%.}
\begin{tabular}{ll}
\begin{tabular}{rrr}
\hline
\multicolumn{3}{c}{Drive Ball-Handler}                                                                     \\ \hline
\multicolumn{1}{c}{Player Name (Team)}            & \multicolumn{1}{l}{FGA}  & \multicolumn{1}{l}{TS\%} \\ \hline
\multicolumn{1}{l}{James Harden (HOU)}    & \multicolumn{1}{r}{1617} & 58.4                               \\
\multicolumn{1}{l}{Darren Collison (SAC)} & \multicolumn{1}{r}{776}  & 57.8                               \\
\multicolumn{1}{l}{LeBron James (CLE)}    & \multicolumn{1}{r}{1416} & 56.8                               \\
\multicolumn{1}{l}{Chris Paul (LAC)}      & \multicolumn{1}{r}{1114} & 56.0                               \\
\multicolumn{1}{l}{Ramon Sessions (WAS)}  & \multicolumn{1}{r}{592}  & 55.4                               \\ \hline
\end{tabular}

&

\begin{tabular}{lrr}
\hline
\multicolumn{3}{c}{Stretch Four}                                                                          \\ \hline
\multicolumn{1}{c}{Player Name (Team)}            & \multicolumn{1}{l}{FGA} & \multicolumn{1}{l}{{TS\%}} \\ \hline
\multicolumn{1}{l}{Jared Dudley (WAS)}    & \multicolumn{1}{r}{487} & 60.2                               \\
\multicolumn{1}{l}{Nemanja Bjelica (MIN)} & \multicolumn{1}{r}{235} & 59.4                               \\
\multicolumn{1}{l}{Marvin Williams (CHA)} & \multicolumn{1}{r}{747} & 58.5                               \\
\multicolumn{1}{l}{Mirza Teletović (PHX)} & \multicolumn{1}{r}{777} & 57.1                               \\
\multicolumn{1}{l}{Nikola Mirotić (CHI)}  & \multicolumn{1}{r}{609} & 56.4                               \\ \hline

\end{tabular}
\end{tabular}
\end{table}

\begin{table}[H]
\centering
\caption{Top 5 players for Corner Shooter and Pure Shooter in TS\%.}
\begin{tabular}{ll}
\begin{tabular}{rrr}
\hline
\multicolumn{3}{c}{Corner Shooter}                                                                          \\ \hline
\multicolumn{1}{c}{Player Name (Team)}              & \multicolumn{1}{l}{FGA} & \multicolumn{1}{l}{TS\%} \\ \hline
\multicolumn{1}{l}{Richard Jefferson (CLE)} & \multicolumn{1}{r}{312} & 58.5                               \\
\multicolumn{1}{l}{Joe Ingles (UTA)}        & \multicolumn{1}{r}{291} & 57.2                               \\
\multicolumn{1}{l}{Jae Crowder (BOS)}       & \multicolumn{1}{r}{813} & 56.5                               \\
\multicolumn{1}{l}{Brandon Rush (UTA)}      & \multicolumn{1}{r}{260} & 56.0                               \\
\multicolumn{1}{l}{Patrick Beverley (HOU)}  & \multicolumn{1}{r}{594} & 55.3                               \\ \hline
\end{tabular}

&

\begin{tabular}{lrr}
\hline
\multicolumn{3}{c}{Pure Shooter}                                                                          \\ \hline
\multicolumn{1}{c}{Player Name (Team)}            & \multicolumn{1}{l}{FGA} & \multicolumn{1}{l}{TS\%} \\ \hline
\multicolumn{1}{l}{J.J. Redick (LAC)}    & \multicolumn{1}{r}{880} & 63.2                               \\
\multicolumn{1}{l}{Kyle Korver (ATL)} & \multicolumn{1}{r}{616} & 57.8                               \\
\multicolumn{1}{l}{Allen Crabbe (POR)} & \multicolumn{1}{r}{678} & 57.2                               \\
\multicolumn{1}{l}{José Calderón (DET)} & \multicolumn{1}{r}{455} & 57.1                               \\
\multicolumn{1}{l}{Eric Gordon (NOP)}  & \multicolumn{1}{r}{552} & 56.5                               \\ \hline

\end{tabular}
\end{tabular}
\end{table}

\newpage
\section{Supplementary Figures}
\begin{figure}[htbp]
    \centering
    \includegraphics[width=0.75\linewidth]{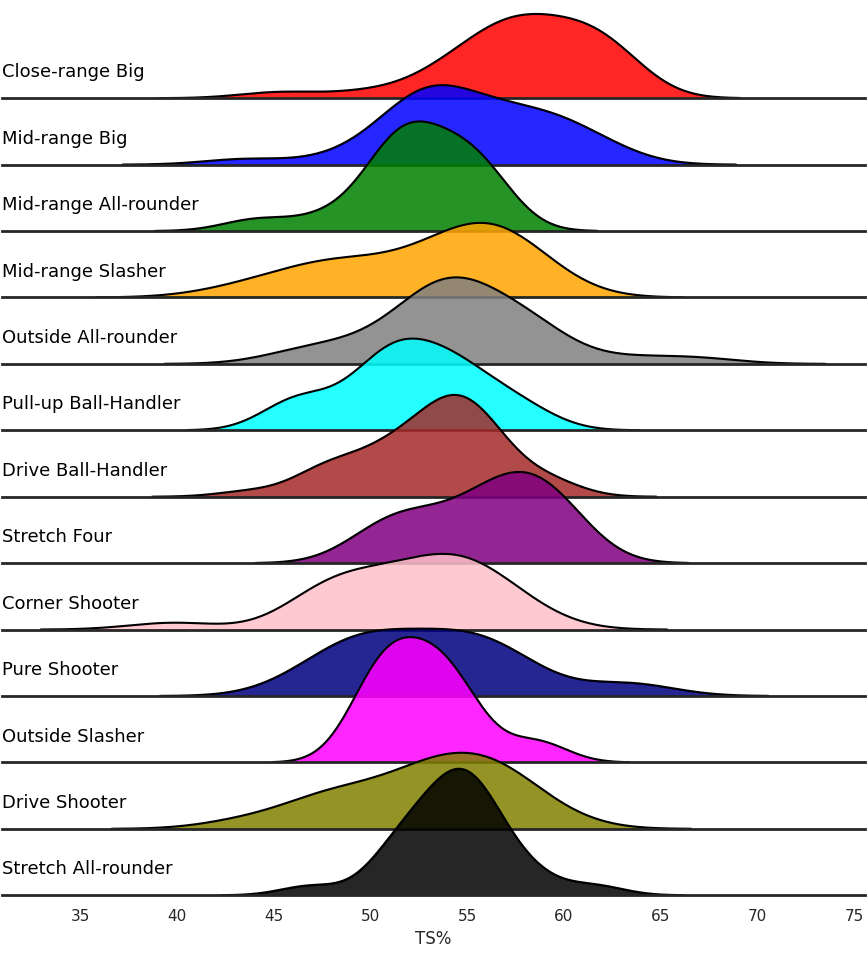}	
    \caption{Ridgeline plot of TS\% by shooting style cluster}
    \label{fig:Ridgeline Plot1}
\end{figure}

\textbf{Comparison of Shooting Style Cluster in Scoring Efficiency.}\\
The scoring efficiencies of each cluster were compared by TS\%. Figure \ref{fig:Ridgeline Plot1} shows a ridgeline plot of the TS\% for each cluster and it can be seen that Close-range Big is scoring efficiently. This may be attributed to the cluster's strong preference for shots from close range.

\newpage
\textbf{Comparison of Offensive Role Cluster in Scoring Efficiency.}\\
Points Per Possession (PPP) was used to compare the scoring efficiency of the offensive role clusters. PPP of an individual player is calculated by the following formula:

\begin{align}
    PPP=\frac{PTS}{FGA+0.44 \times FTA+TO},
\end{align}
where TO is the total number of turnovers. It differs from TS\% in that it takes turnover into account. This indicator was used because playtypes are recorded even if the possession ends in a turnover. In this method, we created ridgeline plots for each cluster and compared the distribution of PPP for the clustered players in the 2022-23 season. We assumed that each player belonged to the cluster with the largest membership coefficient.

\begin{figure}[h]
    \centering
    \includegraphics[width=\linewidth]{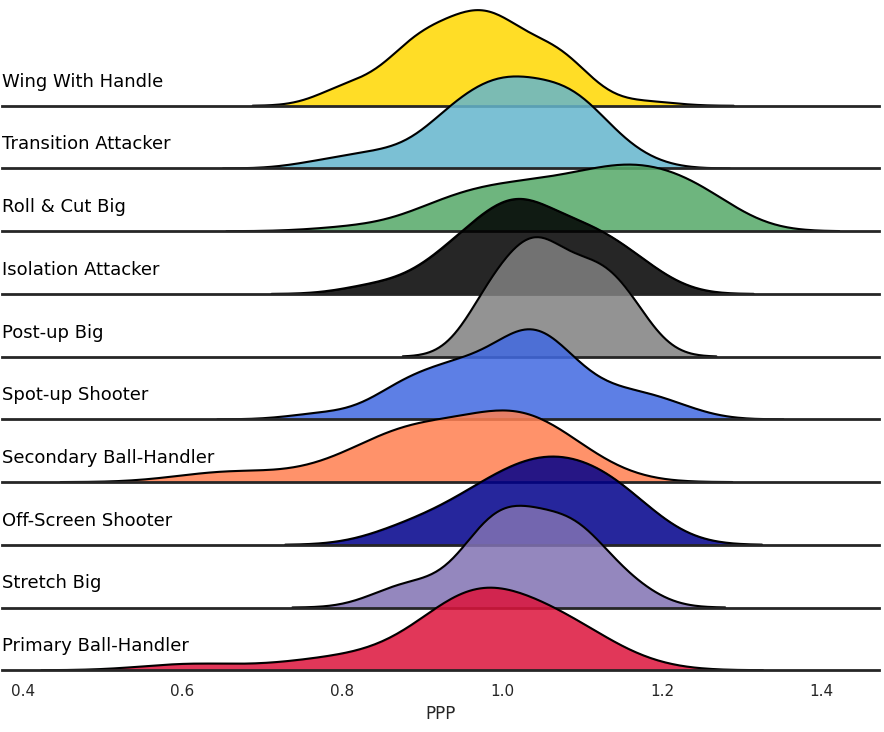}	
    \caption{Ridgeline Plot of PPP by offensive role cluster}
    \label{fig:RidgelinePlot2}
\end{figure}

Figure \ref{fig:RidgelinePlot2} shows the distribution of PPP for each cluster. What can be read from the figure is that Roll \& Cut Big tends to have higher offensive efficiency. This is likely due to the fact that they often dive on the PnR or take offensive rebounds and try to shoot from close range. Secondary Ball-Handler and Primary Ball-Handler players also tend to have slightly lower offensive efficiency. This may be due to their role in initiating the offense, which results in more turnovers, and their tendency to shoot more pull-up jumpers.

\newpage

 
\end{document}